\documentclass{article}

\PassOptionsToPackage{numbers, compress}{natbib}
\usepackage[dblblindworkshop, final]{neurips_2026}
\workshoptitle{Sim2Science: ML with Imperfect Scientific Models}

\usepackage[utf8]{inputenc} 
\usepackage[T1]{fontenc}    
\usepackage{hyperref}       
\usepackage{url}            
\usepackage{booktabs}       
\usepackage{amsfonts}       
\usepackage{amsmath}
\usepackage{float}
\usepackage{graphicx}
\usepackage{nicefrac}       
\usepackage{microtype}      
\usepackage{xcolor}         

\newcommand{\rrBenchN}{640}  

\newcommand{\rrBenchGroups}{7}

\newcommand{\rrTrainNRocks}{368}
\newcommand{\rrTrainNUnits}{51}
\newcommand{\rrValNRocks}{111}
\newcommand{\rrValNUnits}{39}
\newcommand{\rrTestNRocks}{161}
\newcommand{\rrTestNUnits}{45}
\newcommand{\rrBenchNUnits}{135}  
\newcommand{\fsBudgetDraws}{10}  

\newcommand{\rrValZsProd}{13.4}  

\newcommand{\rrValTailedProd}{0.97}
\newcommand{\fsTwentyFiveTailedProd}{3.97}  
\newcommand{\fsRecoveredTailedPct}{76}  

\newcommand{\rrTailedEps}{0.052}
\newcommand{\rrTailedK}{0.507}

\newcommand{\encImputeTestN}{157{,}544}  
\newcommand{\encImputeMajorN}{8}  
\newcommand{\encImputeMajorRSq}{0.89}  
\newcommand{\encImputeTraceN}{13}
\newcommand{\encImputeTraceRSq}{0.84}
\newcommand{\encImputeRSqSi}{0.92}
\newcommand{\encImputeRSqAl}{0.92}
\newcommand{\encImputeRSqCa}{0.88}
\newcommand{\encImputeRSqMg}{0.92}
\newcommand{\encImputeRSqFe}{0.86}
\newcommand{\encImputeRSqK}{0.87}
\newcommand{\encImputeRSqNa}{0.79}
\newcommand{\encImputeRSqTi}{0.90}
\newcommand{\encImputeRSqSr}{0.80}
\newcommand{\encImputeRSqZr}{0.80}
\newcommand{\encImputeRSqBa}{0.77}
\newcommand{\encImputeRSqY}{0.82}
\newcommand{\encImputeRSqRb}{0.89}
\newcommand{\encImputeRSqNi}{0.80}
\newcommand{\encImputeRSqNb}{0.85}
\newcommand{\encImputeRSqV}{0.83}
\newcommand{\encImputeRSqLa}{0.93}
\newcommand{\encImputeRSqCe}{0.94}
\newcommand{\encImputeRSqNd}{0.94}
\newcommand{\encImputeRSqYb}{0.84}
\newcommand{\encImputeRSqTh}{0.88}

\newcommand{\encImputeTraceRSqLinearObj}{0.84}  
\newcommand{\encImputeTraceRSqLogObj}{0.88}  

\newcommand{\fsRocksKnee}{26}  

\newcommand{\fsLatentAmountKnee}{-1.92}

\newcommand{\fsAssayAmountKnee}{-1.47}

\newcommand{\fsAssemblageAmountKnee}{-1.63}

\newcommand{\fsRocksFull}{368}  

\newcommand{\fsLatentAmountFull}{-2.50}

\newcommand{\fsAssayAmountFull}{-2.50}

\newcommand{\fsAssemblageAmountFull}{-2.51}

\newcommand{\fsMaeZeroShotPct}{11.0}  
\newcommand{\fsRecalMaeFullPct}{9.5}  
\newcommand{\fsLatentOverAssayKnee}{0.67}
\newcommand{\fsLatentOverAssayKneeDraws}{9/10}
\newcommand{\fsLatentOverAssayFull}{0.59}
\newcommand{\fsLatentOverAssayFullDraws}{10/10}
\newcommand{\fsLatentOverAssemblageKnee}{0.50}
\newcommand{\fsLatentOverAssemblageKneeDraws}{9/10}
\newcommand{\fsLatentOverAssemblageFull}{0.78}
\newcommand{\fsLatentOverAssemblageFullDraws}{10/10}
\newcommand{\fsLatentOverRecalKnee}{1.38}

\newcommand{\fsLatentOverRecalFull}{0.70}

\newcommand{\fsLatentOverAssayLo}{0.50}  
\newcommand{\fsLatentOverAssayHi}{0.67}

\newcommand{\fsRecalMaeBeatsZeroShotAt}{54}  
\newcommand{\fsRecalMaeBeatsZeroShotPct}{10.6}  
\newcommand{\fsRecalMaeUnanimousAt}{210}  

\newcommand{\icFitRocks}{25}  

\newcommand{\icNSamples}{2{,}048}  
\newcommand{\icRockSilica}{79.9}  
\newcommand{\icTestZsJoint}{9.36}  
\newcommand{\icTestAdJoint}{1.58}

\newcommand{\icDirPct}{71}  
\newcommand{\icOpqRatePct}{93}  
\newcommand{\icBiPZs}{0.51}
\newcommand{\icBiPAd}{0.98}
\newcommand{\icOpqPZs}{0.65}
\newcommand{\icOpqPAd}{0.94}
\newcommand{\icBiShareZsPct}{0.7}
\newcommand{\icBiShareAdPct}{7.5}
\newcommand{\icBiShareMeasPct}{6.0}
\newcommand{\icQShareZsPct}{63.6}
\newcommand{\icQShareAdPct}{51.7}
\newcommand{\icQShareMeasPct}{52.2}

\newcommand{\acCondSamples}{512}  
\newcommand{\acCompDraws}{2{,}048}  

\newcommand{\pvPlag}{79.3}
\newcommand{\pvAfs}{20.0}
\newcommand{\pvCpx}{76.5}
\newcommand{\pvSpl}{51.9}
\newcommand{\pvAmp}{53.3}
\newcommand{\pvIlm}{45.6}
\newcommand{\pvOpx}{25.4}
\newcommand{\pvQ}{34.4}
\newcommand{\pvBi}{27.0}
\newcommand{\pvOl}{10.6}
\newcommand{\pvG}{10.8}
\newcommand{\pvRu}{12.7}
\newcommand{\pvSph}{10.8}
\newcommand{\pvSill}{3.1}
\newcommand{\pvCd}{3.4}
\newcommand{\pvMu}{4.9}
\newcommand{\pvEp}{4.4}
\newcommand{\pvNe}{1.7}
\newcommand{\pvKy}{1.0}
\newcommand{\pvAnd}{0.6}
\newcommand{\pvCor}{0.2}

\newcommand{\rrAurocFsp}{0.870}
\newcommand{\rrSpearmanFsp}{0.456}
\newcommand{\rrNQuantFsp}{601}  
\newcommand{\rrAurocPx}{0.780}
\newcommand{\rrSpearmanPx}{0.396}
\newcommand{\rrNQuantPx}{292}  
\newcommand{\rrAurocAmp}{0.575}
\newcommand{\rrSpearmanAmp}{0.350}
\newcommand{\rrNQuantAmp}{312}  
\newcommand{\rrAurocBi}{0.770}
\newcommand{\rrSpearmanBi}{-0.044}
\newcommand{\rrNQuantBi}{322}  
\newcommand{\rrAurocOl}{0.829}
\newcommand{\rrSpearmanOl}{0.762}
\newcommand{\rrNQuantOl}{79}  
\newcommand{\rrAurocOpq}{0.638}
\newcommand{\rrSpearmanOpq}{0.219}
\newcommand{\rrNQuantOpq}{559}  
\newcommand{\rrAurocQ}{0.853}
\newcommand{\rrSpearmanQ}{0.667}
\newcommand{\rrNQuantQ}{148}  
\newcommand{\rrSpearmanPosGroups}{6}  
\newcommand{\rrOffVocabPctProd}{0.12}  
\newcommand{\sysPhaseVocab}{21}  
\newcommand{\sysRealGroups}{7}  

\newcommand{\sysInputTokens}{22}  
\newcommand{\sysEncoderTokenSlots}{23}  
\newcommand{\sysFewshotBudgetKnee}{25}  
\newcommand{\sysRecalibEpochs}{400}  
\newcommand{\sysPretrainBulksAllrock}{787{,}721}  
\newcommand{\sysCorpusTrainBulksRealized}{9{,}946}
\newcommand{\sysCorpusTrainDraws}{48}  
\newcommand{\sysCorpusTrainRows}{949{,}228}  
\newcommand{\sysCorpusValtestDraws}{128}

\newcommand{\sysCorpusValBulks}{218}  
\newcommand{\sysCorpusValRows}{55{,}576}
\newcommand{\sysCorpusTestBulks}{274}  
\newcommand{\sysCorpusTestRows}{69{,}521}

\newcommand{\sysCompOxides}{11}  
\newcommand{\hdCompEps}{$10^{-4}$}  
\newcommand{\hdModeHidden}{512}
\newcommand{\hdModeLayers}{3}
\newcommand{\hdCompHidden}{128}  
\newcommand{\hdCompLayers}{1}
\newcommand{\hdCondHidden}{512}
\newcommand{\hdCondLayers}{2}
\newcommand{\hdDropout}{0.260}
\newcommand{\hdCondComponents}{3}  
\newcommand{\hdMagWeight}{0.0281}
\newcommand{\hdEffLambda}{0.59}  
\newcommand{\hdLambdaComp}{1.0}
\newcommand{\hdLambdaCond}{0.01}  
\newcommand{\hdLambdaBalance}{$3.08\times10^{-3}$}
\newcommand{\hdLr}{$4.70\times10^{-4}$}
\newcommand{\hdWeightDecay}{$2.46\times10^{-5}$}
\newcommand{\hdBatchSize}{1024}
\newcommand{\hdEpochsCeiling}{120}  
\newcommand{\hdPatience}{10}  
\newcommand{\hdSeed}{0}
\newcommand{\hdEpochsRun}{66}  
\newcommand{\hdModeBestEpoch}{6}
\newcommand{\hdCompBestEpoch}{56}
\newcommand{\hdCondBestEpoch}{42}

\newcommand{\hpoTrials}{48}  
\newcommand{\hpoStartupTrials}{12}  
\newcommand{\hpoTrialEpochs}{60}  
\newcommand{\hpoTieBand}{0.03}  

\newcommand{\hpoRangeLr}{$10^{-4}$ to $10^{-2}$}  
\newcommand{\hpoRangeWeightDecay}{$10^{-7}$ to $10^{-2}$}  
\newcommand{\hpoRangeMagWeight}{0.0048 to 0.048}  
\newcommand{\hpoRangeLambdaBalance}{$10^{-3}$ to $10^{4}$}  
\newcommand{\hpoRangeDropout}{0 to 0.4}  
\newcommand{\hpoRangeModeHidden}{64, 128, 256, 512}  
\newcommand{\hpoRangeModeLayers}{1 to 3}  

\newcommand{\rcFitLr}{0.05}  
\newcommand{\rcWdGridLo}{$10^{-4}$}
\newcommand{\rcWdGridHi}{10}

\newcommand{\sysModeHeadParams}{1{,}346{,}090}
\newcommand{\sysCompHeadParams}{65{,}404}
\newcommand{\sysCondHeadParams}{418{,}843}

\newcommand{\sysEncoderParams}{3{,}234{,}699}  
\newcommand{\sysEncoderDModel}{256}

\newcommand{\sysRecalibVectorParams}{23}
\newcommand{\sysRecalibMatrixParams}{121}
\newcommand{\sysRecalibParams}{123}  
\newcommand{\sysResupervisedParams}{34{,}839}  
\newcommand{\sysResupervisedAssayParams}{4{,}887}  

\newcommand{\rsHidden}{64}  
\newcommand{\rsComponents}{2}  
\newcommand{\rsLr}{0.03}  
\newcommand{\rsEpochs}{1500}  
\newcommand{\rsWeightDecay}{1.0}  
\newcommand{\rsBudgetGrid}{25, 50, 100, 200, and 400}  

\newcommand{\rrOverPctFsp}{72}  
\newcommand{\rrMagFoldFsp}{1.14}  
\newcommand{\rrOverPctPx}{37}
\newcommand{\rrMagFoldPx}{0.76}
\newcommand{\rrOverPctAmp}{46}
\newcommand{\rrMagFoldAmp}{0.92}
\newcommand{\rrOverPctBi}{29}
\newcommand{\rrMagFoldBi}{0.61}
\newcommand{\rrOverPctOl}{41}
\newcommand{\rrMagFoldOl}{0.88}
\newcommand{\rrOverPctOpq}{6}
\newcommand{\rrMagFoldOpq}{0.34}
\newcommand{\rrOverPctQ}{84}
\newcommand{\rrMagFoldQ}{2.01}

\title{Simulated petrological labels enable probabilistic mineralogy prediction from whole-rock geochemistry}

\author{%
  Matthew J. Liu \\
  Materials Sciences Division\\
  Lawrence Berkeley National Laboratory\\
  Berkeley, CA, USA \\
  \texttt{matthewliu@berkeley.edu} \\
  \And
  Aaron D. Kaplan \\
  Materials Sciences Division\\
  Lawrence Berkeley National Laboratory\\
  Berkeley, CA, USA \\
  \texttt{aaron.kaplan.physics@gmail.com} \\
  \And
  Michael L. Whittaker \\
  Energy Geosciences Division\\
  Lawrence Berkeley National Laboratory\\
  Berkeley, CA, USA \\
  \texttt{MWhittaker@lbl.gov} \\
  \And
  Kristin A. Persson \\
  Materials Sciences Division\\
  Lawrence Berkeley National Laboratory\\
  Berkeley, CA, USA \\
  \texttt{kapersson@lbl.gov} \\
}

\begin{document}

\maketitle

\begin{abstract}
  Modal mineralogy, the mineral identities and abundances in rocks, guides geological studies and industrial resource extraction. While measuring modal mineralogy is time-intensive, whole-rock chemical measurements furnish elemental concentrations at higher throughput. Predicting mineralogy from elemental data, however, poses a one-to-many inverse problem, and labeled mineralogical datasets are too sparse to train on at scale. We address both challenges by simulating petrological labels that express multiple formation conditions per composition and at a scale unattainable by measurements. Nearly one million simulated labels supervise the middle component of a three-part model: a transformer encoder self-supervised on roughly 788{,}000 whole-rock compositions, the simulation-trained heads that predict distributions over mineral assemblages, and a recalibration layer that aligns the predictions with measured mineralogy. Zero-shot on \rrBenchN{} volcanic rocks, the model discriminates the presence of most mineral categories and predicts which rocks are richer in each mineral. However, the predicted abundances are systematically offset from measured values. Recalibration fit on as few as \sysFewshotBudgetKnee{} labeled rocks recovers \fsRecoveredTailedPct{} percent of the likelihood improvement from fitting on the full \rrTrainNRocks{}-rock fitting set, while abundance error responds more slowly (\fsMaeZeroShotPct{}\% zero-shot and \fsRecalMaeFullPct{}\% with the full fitting set). This work demonstrates how a machine learning model in the geochemical sciences can both leverage an imperfect simulator and mitigate its systematic biases.\footnote{Code is available at \url{https://github.com/matthew-j-liu/phasecast}.}
\end{abstract}


\section{Introduction}
\label{sec:intro}
Modal mineralogy underpins both geological investigations and the mining value chain. Mineral assemblages reflect the evolution of magmas and inform flowsheet design and extraction methods for critical materials \cite{shi_mineralml_2026, lotter_business_2018, rasool_mineralogical_2025}. Determining modal mineralogy typically requires an expert to count mineral grains under a microscope or perform X-ray diffraction on carefully prepared samples. In contrast, higher-throughput techniques such as inductively coupled plasma mass spectrometry measure whole-rock chemistry at lower cost \cite{van_acker_inductively_2023}. Predicting mineralogy from geochemistry would therefore bridge accessible measurements to richer geological information. However, predicting mineralogy directly from geochemistry is limited by two key obstacles. The first challenge is the paucity of labeled data. While whole-rock geochemistry is routinely compiled in databases (e.g., \texttt{PetDB} and \texttt{GEOROC}) \cite{lehnert_global_2000, sarbas_georoc_2008}, modal mineralogy rarely accompanies these compositions. The second challenge is the one-to-many mapping between elemental composition and the mineral assemblages that could produce it. For a fixed composition, the identity and proportion of phases in an assemblage depend on magmatic temperature, pressure, and water content; kinetics of crystal nucleation, growth, and chemical diffusion; and transport and separation processes such as exsolution. 

We recognized an opportunity to address both obstacles of mineral prediction by leveraging thermodynamic simulators. Solvers such as \texttt{Perple\_X}, \texttt{THERMO-CALC}, and \texttt{MELTS} approximate mineral formation by solving for thermodynamically stable assemblages \cite{ghiorso_chemical_1995, andersson_thermo-calc_2002, connolly_computation_2005}. These simulators minimize Gibbs free energy to obtain the equilibrium assemblage under fixed formation conditions (e.g., temperature, pressure, water content). While exact formation conditions cannot be recovered from a rock, recent solvers such as \texttt{MAGEMin} are fast enough to sweep candidate conditions for every composition \cite{riel_magemin_2022}, enabling diverse label generation. Simulation at scale thus provides the labels that measurements rarely do, and expresses the one-to-many mapping of our inverse problem. The tradeoff, however, is that labels inherit the simulator's imperfections. For example, kinetics and transport are omitted, and formation conditions are assumed, leading to systematic deviation of simulated from measured mineralogy. Even so, we reasoned that simulated petrological labels could supervise a machine learning model, with recalibration on a small set of measured rocks offering a means to mitigate the simulation's systematic errors. In this work, we simulate equilibrium crystallization of volcanic rocks, where the crystals that form remain in the cooling melt. To interrogate our hypothesis, we pretrain a transformer encoder on roughly 788{,}000 whole-rock compositions, train probabilistic heads on the roughly 950{,}000 assemblages simulated across sampled formation conditions, and evaluate the predictions on \rrBenchN{} volcanic rocks before and after recalibration. Appendix~\ref{sec:related_works_appendix} contextualizes our approach among prior work in computational geoscience.


\section{Overview of model architecture and evaluation}
\label{sec:methods}
The model has three components: a pretrained encoder maps a composition to a latent representation; probabilistic heads, trained on \texttt{MAGEMin}-generated labels, predict from the latent representation which minerals are present and in what proportions, each mineral's oxide composition, and the pressure, temperature, and water content of the melt; and an optional recalibration layer fit on measured mineralogy adjusts systematic errors (Appendix~\ref{sec:implementation_appendix}). Note the input is a fixed vector of 9 major and 13 trace element concentrations, as well as an Fe$^{3+}$/total-Fe row; formation conditions are not inputs.

\paragraph{Pretrained composition encoder.}

We adapt the CrabNet transformer architecture, which represents chemical formulas \cite{wang_compositionally_2021}. Each element is one token; self-attention across tokens and a masked mean pool yield one latent vector per rock. We pretrain the encoder by masked-element modeling on \sysPretrainBulksAllrock{} measured whole-rock compositions, filtered from a global geochemical compilation \cite{gard_global_2019}. We replace a random subset of each composition's reported element rows with a learned mask token. A decoder on the pooled latent representation recovers concentrations as standardized log values. Appendices~\ref{sec:tokenization_appendix} -~\ref{sec:pretrain_appendix} detail the encoder and its pretraining, with losses in Fig.~\ref{fig:pretrain_loss}.

\paragraph{Probabilistic heads.}

On the frozen encoder we train three prediction heads supervised by \texttt{MAGEMin}-generated labels. \sysCorpusTrainBulksRealized{} measured compositions, drawn from the same compilation but held out from pretraining itself, are each simulated under \sysCorpusTrainDraws{} sampled formation conditions (referred to as ``draws'', Appendix~\ref{sec:corpus_appendix}). Thus, one composition maps to many possible assemblages, and the heads predict a distribution rather than a single value. The assemblage head predicts over $J = \sysPhaseVocab{}$ mineral phases (Table~\ref{tab:phase_vocab} lists all phases) in two factors. A Bernoulli factor decides which minerals are present, and a Dirichlet factor, conditioned on present minerals, predicts their abundances:

\begin{equation}\label{eq:hurdle}
p(s, m \mid h) = \underbrace{\prod_{j=1}^{J} p_j^{s_j}\,(1 - p_j)^{1 - s_j}}_{\text{presence}}
\;\cdot\;
\underbrace{\mathrm{Dir}\big(m_s \,;\, \alpha_s(h, s)\big)}_{\text{abundance given presence}}
\end{equation}

with $s$ the presence vector, $m$ the abundances, $p_j$ the presence probability of phase $j$, and $\alpha_s(h, s)$ the concentrations of the phases $s$ predicts. At training $s$ is the true support; at inference it is drawn from the presence factor. Appendices~\ref{sec:heads_appendix} and~\ref{sec:training_appendix} detail the other two heads and the joint training, with per-head losses in Fig.~\ref{fig:heads_loss}.

\paragraph{Recalibration layer.}

We employ post-hoc recalibration as a parameter-efficient approach for adapting the simulation-trained heads to real rocks. The recalibration layer learns from a small number of labeled rocks while leaving the pretrained encoder and prediction heads fixed, enabling few-shot adaptation \cite{guo_calibration_2017, kuleshov_accurate_2018}. The layer corrects each factor of Eq.~\ref{eq:hurdle} by an affine map on the presence logits $\ell$ and a multiplicative rescaling of the Dirichlet concentrations $\alpha$:

\begin{equation}\label{eq:recal}
\ell' = a \odot \ell + b + A^{\top}\ell,
\qquad
\alpha' = \alpha \odot \exp\big(c + B^{\top} \log \alpha\big)
\end{equation}

where $\odot$ is the element-wise product and $(a, b, c, A, B)$ are the learned parameters (Appendix~\ref{sec:recalibrator_appendix}).

\paragraph{Modal mineralogy evaluation.}

We evaluate the model on \rrBenchN{} volcanic rocks from two studies by the U.S. Geological Survey reporting whole-rock chemistry and mineral abundances \cite{du_bray_petrographic_2016, du_bray_petrographic_2019}. Rocks within a stratigraphic unit possess similar chemistry, motivating data splits by these units to prevent evaluating rocks from the same groups used for training. Appendix~\ref{sec:benchmark_appendix} details how candidate train/validation/test splits were generated and filtered. The resulting train, validation, and test sets contain \rrTrainNRocks{}, \rrValNRocks{}, and \rrTestNRocks{} rocks.

The sources report \rrBenchGroups{} mineral categories, onto which 10 of the heads' \sysPhaseVocab{} phases map (Table \ref{tab:magemin_collapse}). We allowed the model to predict the other 11 phases rather than impose their absence, using AUROC to assess presence discrimination (Appendix~\ref{sec:discrimination_appendix}). Predictions are scored by joint negative log-likelihood (NLL) of the measured assemblage under Eq.~\ref{eq:hurdle}, with the predicted distribution collapsed onto the \rrBenchGroups{} mineral categories. The NLL is reported in nats, where a lower score indicates that the predicted distribution assigns higher probability density to the measured mineralogy (i.e., model confidence). We complement the likelihood with the absolute error between the measured shares and the mean of the predicted share distribution (i.e., accuracy). Because the benchmark reports modal mineralogy alone, we evaluate only the assemblage head. The composition and conditions heads nonetheless predict oxide compositions and formation conditions for every rock (Appendix~\ref{sec:compcond_appendix}).


\section{Results and discussion}
\label{sec:results}

\paragraph{Pretrained representation outperforms raw geochemistry at presence predictions.}
Given that predictions of modal mineralogy depend on the latent representation learned by the encoder, we begin by quantifying how well the representation encodes whole-rock geochemistry. We perform imputation, where one reported element is hidden from a composition and a decoder is trained to recover its concentration from the frozen latent representation. Accurate imputation implies that the representation has learned geochemical structure where elemental concentrations co-vary with factors such as tectonic setting. Table~\ref{tab:imputation_per_element} shows that the decoder recovers elemental concentration at a median log-space $R^2$ of \encImputeMajorRSq{} across \encImputeMajorN{} major elements and \encImputeTraceRSq{} across \encImputeTraceN{} trace elements, scored on \encImputeTestN{} held-out compositions. Thus, the latent representation preserves the geochemistry of rocks.

To interrogate whether the latent representation improves prediction of modal mineralogy, we trained two heads (i.e., multilayer perceptrons) of the form in Eq.~\ref{eq:hurdle} on benchmark mineral abundances. The heads were identical except for their inputs: either raw geochemistry or the latent representation. Fits ranged from \sysFewshotBudgetKnee{} rocks to the full \fsRocksFull{}-rock fitting set, with \fsBudgetDraws{} replicates performed at each size. At every size, the head given the latent representation outperformed the head given raw geochemistry by \fsLatentOverAssayLo{} to \fsLatentOverAssayHi{} nats of joint NLL (paired within replicates). Fig.~\ref{fig:fewshot_decomp} decomposes the difference, revealing that the presence factor is the primary source of the advantage. A third head given no chemistry converges to the same abundance factor (Appendix~\ref{sec:resupervision_appendix}), suggesting that abundance is learned from the minerals present rather than from the chemical inputs. Pretraining therefore improves mineralogy predictions by improving discernment of mineral presence.

\begin{figure}[h]
    \centering
    \includegraphics[width=0.80\linewidth]{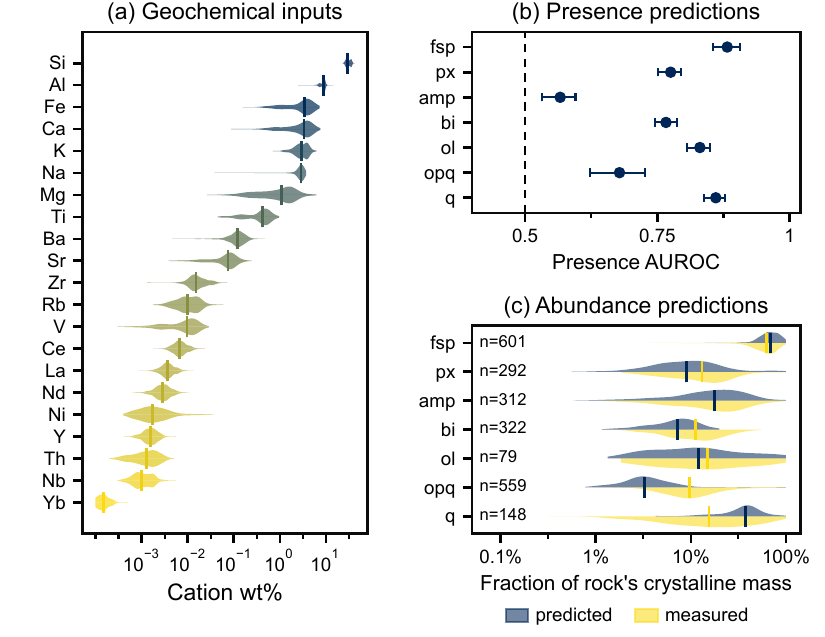} 
    \caption{\textbf{Zero-shot predictions over \rrBenchN{} volcanic rocks.} \textbf{(a)} Model inputs. \textbf{(b)} Presence AUROC per mineral category, where the dotted line represents a random classifier. Whiskers span the 25th to 75th percentile across bootstrap resamples of stratigraphic units. \textbf{(c)} Violin plots for predicted vs. measured mineral abundance when the heads are given the measured mineral list as their support (i.e., teacher-forced). In (a) and (c), violin width reflects the density of rocks at each value and the tick marks the median. Mineral abbreviations in Table \ref{tab:magemin_collapse}.}
    \label{fig:cohort_card}
\end{figure}

\paragraph{Zero-shot predictions recover mineral ordering but misestimate abundance.} 

While the baseline heads are in principle able to predict modal mineralogy, their predictions are confined to one set of minerals. To develop a more robust predictor we supervise the heads with simulated labels from \texttt{MAGEMin}. The simulated labels widen the mineral set to \sysPhaseVocab{} phases where any of the $2^{\sysPhaseVocab}$ subsets can be a predicted assemblage. Because the heads learn from simulated labels alone, their predictions on the benchmark are zero-shot. Fig.~\ref{fig:cohort_card}a pairs the whole-rock compositions entering the model with their predicted presence (Fig.~\ref{fig:cohort_card}b) and abundance (Fig.~\ref{fig:cohort_card}c). Rocks assigned higher predicted presence probabilities for a mineral category contain that mineral more often (AUROC of \rrAurocAmp{} to \rrAurocFsp{} across categories; Table~\ref{tab:percat_auroc}). Of the 7 mineral categories, amphibole exhibits the lowest AUROC, consistent with the strong dependence between amphibole formation and dissolved water content \cite{krawczynski_amphibole_2012} (Fig.~\ref{fig:amp_water_dose}). The 11 phases outside the benchmark categories are together predicted at a mean \rrOffVocabPctProd{} wt\%, an absence learned by the model rather than imposed. When a mineral category is present, the predicted abundance and measured abundance are positively correlated (Spearman rank correlation positive for \rrSpearmanPosGroups{} of \rrBenchGroups{} categories, Table~\ref{tab:percat_spearman}), though predicted abundances exhibit category-specific bias (Fig.~\ref{fig:cohort_card}c). For example, feldspar and quartz are over-predicted the most often (Table~\ref{tab:percat_bias}). Given that quartz subsumes its polymorphs (tridymite and cristobalite) and feldspar spans solid solutions, we hypothesize that composition alone under-determines these families. Nonetheless, the zero-shot model ranks minerals and rocks in the measured order, with systematic bias on abundance.

\paragraph{Recalibration restores confidence with few measured labels.}
\label{sec:results_recovery}

\begin{figure}[h]
    \centering
    \includegraphics[width=0.80\linewidth]{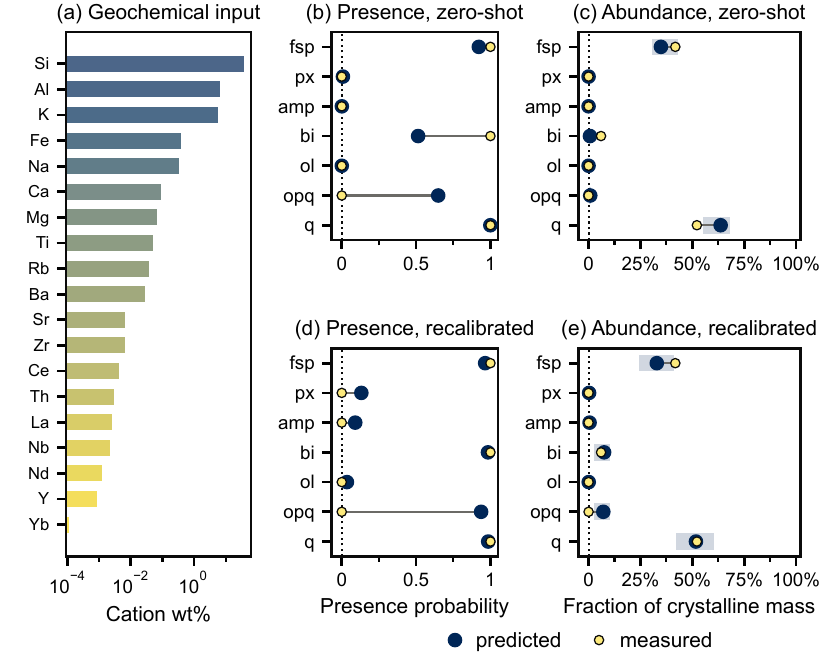}
    \caption{\textbf{Recalibration improves mineral predictions.} Zero-shot vs. recalibrated predictions on the test set rock with the median recalibrated joint NLL. \textbf{(a)} Model input. \textbf{(b, c)} Zero-shot presence probability and mineral abundance. \textbf{(d, e)} Analogous predictions after a \icFitRocks-rock fit. Abundance panels (c and e) each summarize \icNSamples{} candidate assemblages sampled from the presence probabilities of (b and d), respectively. Points denote the mean across candidates, shaded bands indicate the 25th to 75th percentile, and gray lines connect predicted to measured values. Mineral abbreviations in Table \ref{tab:magemin_collapse}.}
    \label{fig:inference_card}
\end{figure}

To address the systematic error in the zero-shot predictions, we employ post-hoc recalibration, fitting the layer of Eq.~\ref{eq:recal} on labeled rocks to align predicted distributions with measured mineralogy \cite{guo_calibration_2017, kuleshov_accurate_2018}. Fitting on \icFitRocks{} rocks reduced the joint NLL on the test set from \icTestZsJoint{} to \icTestAdJoint{} nats, with \icDirPct{} percent of the reduction in the abundance factor. Fig.~\ref{fig:inference_card} illustrates the correction on one rhyolite from the reserved test set (\icRockSilica{} wt\% SiO$_2$, Fig.~\ref{fig:inference_card}a) after a fit on \icFitRocks{} rocks. The rhyolite rock was selected because it had the median recalibrated joint NLL, making it representative rather than curated. The zero-shot model assigns biotite (present in the rock) a presence probability of \icBiPZs{}, while assigning \icOpqPZs{} to the iron-titanium oxides (absent from the rock, Fig.~\ref{fig:inference_card}b). In the zero-shot abundances (Fig.~\ref{fig:inference_card}c), the predicted quartz abundance overshoots at \icQShareZsPct{}\% compared to \icQShareMeasPct{}\% measured while biotite undershoots at \icBiShareZsPct{}\% compared to \icBiShareMeasPct{}\%. Recalibration correctly raises the predicted probability for biotite to \icBiPAd{} (Fig.~\ref{fig:inference_card}d) and brings the recalibrated abundance predictions for both quartz and biotite (\icQShareAdPct{}\% and \icBiShareAdPct{}\%, respectively) toward their measured values (Fig.~\ref{fig:inference_card}e). However, the recalibration also raises the iron-titanium oxides probability to \icOpqPAd{}, likely reflecting the \icOpqRatePct{}\% rate at which the category appears across the test set.

Lastly, we examine how recalibration performance -- with respect to both model confidence (NLL) and accuracy (MAE) -- scales with fit sizes. Comparisons are evaluated on the validation set, having reserved the test set for the single fit of Fig.~\ref{fig:inference_card}. Across the benchmark, fitting the recalibration layer on the full \rrTrainNRocks-rock fitting set reduced validation joint NLL from \rrValZsProd{} to \rrValTailedProd{} nats. Notably, fits of about \sysFewshotBudgetKnee{} rocks recovered \fsRecoveredTailedPct{} percent of the reduction, achieving \fsTwentyFiveTailedProd{} nats (Appendix~\ref{sec:resupervision_appendix}). Accuracy improves more slowly over the same fits. With no rocks fit, predicted abundances are off by \fsMaeZeroShotPct{}\%, decreasing to \fsRecalMaeBeatsZeroShotPct{}\% at \fsRecalMaeBeatsZeroShotAt{} rocks and \fsRecalMaeFullPct{}\% on the full fitting set. The \sysFewshotBudgetKnee{} rocks that recover most of the confidence therefore recover comparatively less accuracy. The foregoing fits also facilitate comparison of simulated and measured mineralogy as label sources. Appendix~\ref{sec:resupervision_appendix} details how a head trained directly on measured abundances reaches the lower joint NLL while the recalibrated model reaches the lower abundance error. The head trained on measured abundances, however, requires the measured mineralogy whose scarcity motivates this work and is restricted to the \rrBenchGroups{} categories its labels resolve; the simulation-trained heads predict a distribution over the $2^{\sysPhaseVocab}$ candidate assemblages, with accompanying predictions of oxide compositions and formation conditions. 

\paragraph{Limitations and conclusion.} The benchmark records phenocryst modes from two adjacent volcanic fields, whereas the labels are equilibrium solids at 50 and 30 wt\% residual melt. The absolute errors are teacher-forced, the fitting-size curves share the validation set used to select regularization, and no external method is evaluated. Composition alone may under-determine polymorphs and solid solutions, where errors concentrate; pretraining on crystal structures could resolve them. More broadly, leveraging different geological simulators, or advancing the simulators themselves, enables re-supervision of the heads without affecting the underlying latent representation. We therefore envision the framework expanding to accommodate new phases and assemblages, with applications across geology and mineral prospecting.


\begin{ack}
M.J.L. is supported by an Arnold O. Beckman Postdoctoral Fellowship in Chemical Sciences from The Arnold and Mabel Beckman Foundation. This work was also supported by the Minerals to Materials Supply Chain Research Facility (METALLIC), funded by the U.S. Department of Energy, Office of Critical Minerals and Energy Innovation (CMEI)

\end{ack}

\bibliography{references}
\newpage

\appendix

\section{Related Works}
\label{sec:related_works_appendix}
\renewcommand{\thetable}{A\arabic{table}}
\renewcommand{\theHtable}{A\arabic{table}}
\setcounter{table}{0}

Machine learning methods have been developed and applied broadly across geoscience, with recent reviews surveying progress in the field \cite{he_review_2022, dramsch_70_2020}. We note that mineral prospecting from geophysical and geospatial data prominently features machine learning methods as well \cite{mou_interpretable_2025, cheng_gis-based_2023, amirajlo_ensemble_2026, nair_masked_2025}, but we limit discussion here to the geochemical sciences. Within that scope, the prediction task involves the two obstacles in Section \ref{sec:intro} of the main text: a one-to-many mapping from composition to assemblage and a scarcity of labeled mineralogy at training scale. Each family of work below resolves at most one of these challenges. We saw the opportunity to address both by building one predictor that spans rock types, enabled by using a thermodynamic simulator.

Tabular, supervised models represent one prominent group of machine learning applications in geochemistry, each fit to a specific dataset obtained at a site of operation (e.g., ore deposit, processing facility). For example, a
neural network trained on 77 samples from a basin well in Sichuan predicts mineral abundances from X-ray fluorescence measurements, reporting $R^2$ of 0.8 or higher for most minerals \cite{ye_machine_2025}. Gradient-boosted and
kernel models have been developed to predict rare earth element concentrations in Tethyan phosphorites \cite{tahar-belkacem_machine_2024}, rare earth enrichment in sedimentary strata \cite{bishop_using_2024}, and rare earth recovery from secondary resources \cite{nguyen_explainable_2025}. Because these models serve their site of operation, their evaluation is often scored on held-out samples from the same site. And while performance is often strong, each new element, mineral, or lithology requires a labeled dataset of its own. Thus, extending one of these models to a new setting requires characterization measurements (e.g., optical point counting or X-ray diffraction) at scale. In the context of the two obstacles to mineralogy prediction, tabular models do not serve as a general predictor.

A second family of computational applications in geochemistry comprises models that use the chemistry of individual mineral grains (spot analyses) as inputs. For example, \texttt{mineralML} and \texttt{Qmin} classify mineral grains from electron-microprobe analyses \cite{shi_mineralml_2026, da_silva_qmin_2021}, while stoichiometric rules accomplish similar identification without learning \cite{siebach_mist_2026}. Beyond mineral identity, single-mineral models predict formation conditions from trace elements (e.g., in quartz) \cite{wang_machine_2021}. These tools provide high accuracy options for identifying mineral presence at the grain scale, but are unable to predict modal abundance at the whole-rock scale. 

Estimating modes from whole-rock geochemistry has remained a challenge in geochemical research and historically has not employed machine learning methods. Normative calculation allocates oxides to an idealized anhydrous mineral set by fixed rules, from the CIPW norm through its successive implementations \cite{cross_quantitative_1902, gonzalez-guzman_shinynorrrm_2023, buckle_webnorm_2023, williams_pyrolite_2020}. These calculations return a single deterministic assemblage for a given bulk composition, independent of the conditions under which the rock formed. Mass-balance and unmixing treatments instead solve for the proportions of a mineral set by least squares \cite{paktunc_modan_1998} or as a posterior over both the mixture coefficients and the endmember compositions \cite{tolosana-delgado_constructing_2011}. The latter allows a probabilistic prediction, but are fit one study at a time and require a mineral list to be defined upfront. 

Thermodynamic simulators in geochemistry have largely remained as calculators, with their output serving as the primary deliverable of a study rather than as a training signal for a model. In contrast, adjacent fields have pursued simulation-to-reality pipelines in various contexts. For example, the materials science community transfers models from first-principles calculations to experiments \cite{yahagi_transfer_2025}, bridges simulation and measurement with uncertainty-aware learning \cite{chen_uncertainty_2026}, and measures scaling laws for simulation-to-real transfer \cite{minami_scaling_2025}. We reasoned that the same paradigm could address the two obstacles to mineralogy prediction. Simulated labels generated at training scale address label scarcity and sampled formation conditions express the one-to-many mapping. 

The family of models we surveyed -- site-specific models, grain-scale classifiers, normative calculators, and simulators -- each address distinct challenges in their respective fields. Table~\ref{tab:related_works} summarizes what our approach adopts from each family of models. We adapt a composition transformer built for property prediction \cite{wang_compositionally_2021} to whole-rock geochemistry (Section~\ref{sec:methods}), and import the masked pretraining objective from geospatial prospecting \cite{nair_masked_2025} to whole-rock geochemistry as well. Because the representation is learned once from unlabeled compositions and the supervision is generated by the simulator, extending the predictor to a new target does not require an extensive labeled dataset assembled per application. Rather, new supervision is regenerated from the simulator and a new head is trained against it. Within the fixed set of candidate phases, the model predicts which minerals are present in each rock and returns its predictions as distributions over assemblages, preserving the one-to-many mapping that a single point estimate instead averages over. Our combined framework also draws from the simulation-to-real literature. The recalibration layer is fit on a small held-out set of measured mineralogy to learn a correction that adjusts systematic errors in the heads' predictions \cite{guo_calibration_2017, kuleshov_accurate_2018}. A correction of this form can account for systematic overconfidence while preserving the model's ranking of minerals. Lastly, an emerging paradigm in transfer learning contends that adapting a pretrained model through a small map on frozen features outperforms training on the target task alone when target labels are scarce \cite{tahir_features_2025}, as we have done in this work. To our knowledge, the combination of self-supervised pretraining, simulated labels, and post-hoc recalibration on measured mineralogy has not been reported in a machine learning model for the geochemical sciences.

\begin{table}
\caption{Prior families of work in geochemical prediction, the challenge each addresses in its respective field, and what this work adopts from each.}
\label{tab:related_works}
\centering
\begin{tabular}{p{0.26\linewidth}p{0.36\linewidth}p{0.28\linewidth}}
  \toprule
  \multicolumn{1}{c}{\textbf{Family}} &
  \multicolumn{1}{c}{\textbf{Challenge addressed}} &
  \multicolumn{1}{c}{\textbf{Adopted in this work}} \\
  \midrule
  Site-specific tabular models &
  Predicting quantities that are costly to measure at an operating site &
  -- \\
  Grain-scale classifiers &
  Automated mineral identification from spot chemistry &
  Stoichiometric naming rules (Appendix~\ref{sec:corpus_appendix}) \\
  Normative and unmixing calculation &
  Estimating mineralogy from bulk composition &
  The prediction task and precedent for probabilistic output \\
  Thermodynamic simulators &
  Computing equilibrium assemblages at fixed conditions &
  The label generator \\
  Simulation-to-real transfer &
  Carrying simulation-trained models to real-world measurements &
  Recalibration paradigm \\
  \bottomrule
\end{tabular}
\end{table}

\newpage

\section{Implementation Details}
\label{sec:implementation_appendix}
\renewcommand{\thetable}{B\arabic{table}}
\renewcommand{\theHtable}{B\arabic{table}}
\setcounter{table}{0}

The following subsections trace the model architecture from end to end. We begin with the tokenization of a whole-rock composition (Appendix~\ref{sec:tokenization_appendix}), which a pretrained transformer encodes into one latent representation per rock (Appendices~\ref{sec:architecture_appendix}--\ref{sec:pretrain_appendix}). The latent representation is fed to probabilistic heads, which are trained on a \texttt{MAGEMin}-generated corpus of petrological labels (Appendices~\ref{sec:corpus_appendix}--\ref{sec:training_appendix}). The heads themselves are capable of making zero-shot predictions on real rocks, or their predictions may feed into a recalibration layer (Appendix~\ref{sec:recalibrator_appendix}). We then detail the assembly of the benchmark, which contains whole-rock compositions with measured mineral proportions (Appendix~\ref{sec:benchmark_appendix}). Lastly, we describe how presence discrimination is evaluated on the benchmark (Appendix~\ref{sec:discrimination_appendix}).

All experiments were run on an Apple M4 Max MacBook Pro with 128 GB RAM. The MPS accelerator was used for model training and CPU for \texttt{MAGEMin} simulation. Generating the simulated label corpus was the largest cost, taking roughly 62 hours for the training split and 10 hours for the validation/test split. Encoder pretraining took roughly 3.5 hours, with its two hyperparameter searches taking an additional 13 hours combined. The joint-heads hyperparameter search ran for roughly 15 hours; subsequent retraining using the selected configurations completed in less than an hour. The recalibration layer and the heads trained on measured abundances fitted in seconds to minutes each. Beyond the reported experiments, roughly 100 hours were used for preliminary iterations focused on generating the simulated label corpus and performing hyperparameter sweeps.

\subsection{Input representation and tokenization}
\label{sec:tokenization_appendix}

We adapted the CrabNet architecture \cite{wang_compositionally_2021} with two changes. Firstly, we fixed the rows to a vocabulary of \sysInputTokens{} elements rather than allowing each sample to determine them. The original CrabNet was used to predict material properties of chemical compounds, which inherently provide every element present in the material. In contrast, a whole-rock composition is measured over a standardized set of elements. Fixing the rows therefore makes a given row denote the same element in every rock. Secondly, we marked unreported elements by a reserved padding value rather than by zero abundance. Our use of pretraining necessitates the latter change because the pretraining procedure hides measured concentrations by zeroing them.

A whole-rock composition enters the encoder as a fixed-length, fixed-order sequence of \sysEncoderTokenSlots{} rows composed of \sysInputTokens{} elements plus one reserved row for the Fe$^{3+}$/total-Fe ratio (Table~\ref{tab:element_vocab}). Of the \sysInputTokens{} elements, nine are major elements derived from oxide weight percent while thirteen are trace elements reported in parts per million. We selected the thirteen trace elements based on two filtering criteria applied to the global geochemical compilation \cite{gard_global_2019}. (1) each element is reported for at least one third of the compilation's rocks, and (2) each element is routinely used to distinguish rock types that share similar major-element chemistry \cite{he_review_2022, williams_pyrolite_2020}.

\begin{table}[h]
\caption{The fixed elemental inputs read by the encoder.}
\label{tab:element_vocab}
\centering
\begin{tabular}{llll}
\toprule
\multicolumn{1}{c}{\textbf{Group}} &
\multicolumn{1}{c}{\textbf{Count}} &
\multicolumn{1}{c}{\textbf{Members}} &
\multicolumn{1}{c}{\textbf{\texttt{frac}}} \\
\midrule
Majors & 9  & Si, Al, Ca, Mg, Fe, K, Na, Ti, Cr & wt\% / 100 \\
Traces & 13 & Sr, Zr, Ba, Y, Rb, Ni, Nb, V, La, Ce, Nd, Yb, Th & ppm / $10^6$ \\
Redox  & 1  & Fe$^{3+}$/total Fe & ratio, unscaled \\
\bottomrule
\end{tabular}
\end{table}

Each of the \sysEncoderTokenSlots{} rows entering the encoder is one token, stored as a pair of numbers \texttt{(src, frac)}, where \texttt{src} identifies the element by its atomic number and \texttt{frac} is the element's abundance rescaled into the unit interval. The conversions in Table~\ref{tab:element_vocab} perform the rescaling, dividing weight percent by 100 for the majors and parts per million by $10^6$ for the traces. A slot in the sequence can also hold no element at all, with Table~\ref{tab:token_states} showing all four states the encoder can carry. (1) A padding slot marks an element left unreported, with \texttt{src = 0}. (2) A populated slot is used for a reported element. (3) A masked slot hides a reported element during pretraining, taking the reserved identity \texttt{src = 119}. (4) The redox slot carries the Fe$^{3+}$/total-Fe ratio under its own identity, \texttt{src = 120}. Masked slots remain attended so the encoder can predict at them; only padding is excluded from attention. Blanks in the global geochemical compilation enter as padding slots, treated as missing rather than as measured zeros. 

\begin{table}[h]
\caption{Possible states for each token of the input sequence.}
\label{tab:token_states}
\centering
\begin{tabular}{llll}
\toprule
\multicolumn{1}{c}{\textbf{State}} &
\multicolumn{1}{c}{\textbf{\texttt{src}}} &
\multicolumn{1}{c}{\textbf{\texttt{frac}}} &
\multicolumn{1}{c}{\textbf{Attended}} \\
\midrule
Padding (element not reported) & 0 & 0 & no \\
Populated & 1--118 & $>0$ & yes \\
Masked for pretraining & 119 & 0 & yes \\
Fe-redox ratio & 120 & ratio in $[0,1]$ & yes \\
\bottomrule
\end{tabular}
\end{table}

\paragraph{Redox row in production.} We note that the redox row is only populated in pretraining. The heads are trained and evaluated with the row padded such that the padded state is a learned signal rather than an average over the unreported ratio. We pad the redox row because none of the real rocks used for benchmarking report both FeO and Fe$_2$O$_3$. The reserved row is nonetheless an intentional choice, given that every simulated label is already generated under an imposed Fe$^{3+}$/total-Fe ratio (Appendix~\ref{sec:corpus_appendix}). In addition, roughly a quarter of the pretraining set reports the pair of iron concentrations the ratio requires. Thus, for future datasets that do report both, heads retrained with the row populated would allow measured oxidation state to inform the predicted assemblage.

\subsection{Encoder architecture}
\label{sec:architecture_appendix}

We employ CrabNet's featurization without additional modification \cite{wang_compositionally_2021}. Each row's token sums a \texttt{mat2vec} element embedding \cite{tshitoyan_unsupervised_2019} with two sinusoidal abundance encodings, one over linearly discretized \texttt{frac} and one over $\log_{10}$ \texttt{frac}. The element embedding table is initialized from the published 200-dimensional embeddings and remains trainable during pretraining, with the padding row pinned to zero and the mask and redox rows initialized as small random values. The transformer stack is our own implementation with Table~\ref{tab:encoder_hparams} detailing the configuration. Of the tabulated values, the model width, layer count, head count and dropout were selected by the hyperparameter searches described in Appendix~\ref{sec:pretrain_appendix}, while the remainder are fixed conventions informed by the original CrabNet implementation.

\begin{table}[h]
\caption{Encoder configuration.}
\label{tab:encoder_hparams}
\centering
\begin{tabular}{lll}
\toprule
\multicolumn{1}{c}{\textbf{Hyperparameter}} &
\multicolumn{1}{c}{\textbf{Meaning}} &
\multicolumn{1}{c}{\textbf{Value}} \\
\midrule
\texttt{d\_model} & Dimensionality of latent representation & 256 \\
\texttt{n\_layers} & Transformer encoder layers & 4 \\
\texttt{n\_heads} & Attention heads per layer & 8 \\
\texttt{dim\_feedforward} & Feed-forward width, $4\times$ \texttt{d\_model} & 1024 \\
\texttt{activation} & Non-linear activation function & GELU \\
\texttt{norm\_first} & Pre-norm residual arrangement & true \\
\texttt{dropout} & Attention and feed-forward dropout & 0.0582 \\
\texttt{frac\_resolution} & Sinusoidal lookup bins per abundance encoding & 5000 \\
Element embedding & Rows $\times$ dimension & $121 \times 200$ \\
Pooling & Mean over non-padding rows & masked mean \\
\bottomrule
\end{tabular}
\end{table}

For a batch of $B$ rocks, \texttt{src} and \texttt{frac} have shape $[B, 23]$, the token grid and the contextualized grid are $[B, 23, 256]$, and the pooled latent representation $h$ is $[B, 256]$. The encoder checkpoint carries 4{,}514{,}699 tensor entries, of which \sysEncoderParams{} are trainable parameters. The trainable count is derived as follows: the element embedding table contains $121 \times 200 = 24{,}200$ parameters, its projection to the model width contains $200 \times 256 + 256 = 51{,}456$, the scalar gains on the embedding and the two abundance encodings contribute 3, and the four transformer layers hold $789{,}760$ each. The remaining $2 \times 5000 \times 128 = 1{,}280{,}000$ entries are the two sinusoidal lookup tables, which are fixed rather than learned.

\subsection{Self-supervised pretraining}
\label{sec:pretrain_appendix}

\paragraph{Pretraining data.} All pretraining data derives from one public compilation of whole-rock geochemistry \cite{gard_global_2019}, which aggregates roughly one million published compositions spanning all rock types. We filtered the compilation to the pretraining set as follows. Oxides above 100 wt\% are treated as aphysical and categorized as missing values (i.e., padding). Rocks with fewer than four populated rows were dropped because masking would leave little context to predict from. Lastly, the compilation contains duplicated compositions under different sample identifiers. We performed deduplication by comparing tokenized composition, dropping any row where \texttt{(src, frac)} values exactly match another row. The result of filtering is \sysPretrainBulksAllrock{} rows for the pretraining set, as referenced in Section~\ref{sec:methods}. 

\paragraph{Held-out construction.} 20\% of the pretraining rows were held out for validation and 20\% for testing, split by row. To ensure that the encoder does not pretrain on compositions used downstream (e.g., as \texttt{MAGEMin} inputs), we assign these compositions to the test split. When a composition assigned to the test split is duplicated under a different sample identifier, deduplication keeps the test-split copy; all other duplicates resolve to the first occurrence.

\paragraph{Masking and objective.} For each rock, a fixed fraction of the populated rows is selected uniformly at random for masking, with a new selection made every epoch. Masked rows receive the mask identity \texttt{src = 119} with \texttt{frac = 0}, and the encoder still attends to them, as discussed in Appendix~\ref{sec:tokenization_appendix}. Writing $M$ for the masked rows of a composition $x$, $\tilde{x}$ for the corrupted composition they produce, and $h_{\tilde{x}}$ for the pooled latent representation, the decoder $g$ regresses the standardized log concentration of each hidden element.

\begin{equation}\label{eq:ssl}
\mathcal{L}_{\mathrm{SSL}} = \mathbb{E}_{x, M}\left[\frac{1}{|M|}
\sum_{i \in M} \big(g_i(h_{\tilde{x}}) - \tilde{y}_i\big)^2\right],
\qquad \tilde{y}_i = \frac{\log_{10}(f_i) - \mathbb{E}_x[\log_{10}(f_i)]}{\sqrt{\mathrm{Var}_x[\log_{10}(f_i)]}}
\end{equation}

where $f_i$ is element $i$'s abundance, and the mean and variance are taken over the pretraining set. The decoder is a two-layer MLP that reads the pooled latent representation. The hidden layer has width \texttt{d\_model}/2 with a GELU activation and the output layer is linear, producing one predicted concentration per row of the input sequence. The output carries no activation function because the regression target, a standardized log concentration, is unbounded. We note that the decoder exists only to train the encoder and is discarded after pretraining.

The loss is computed at masked position such that predictions at unmasked positions receive no gradient. We employed log space for the regression because majors and traces enter \texttt{frac} at magnitudes differing by up to eight orders of magnitude. As a result, a linear-space squared error possesses a negligible gradient for the trace elements. Although log space equalizes the scale of the targets, elements still differ in the variance of their log concentrations. We therefore standardized each element separately. We verified the combined choice on a held-out imputation benchmark, in which each element is predicted from the encoder latent representation and scored by $R^2$ on log concentrations. Relative to the linear-space objective, the median across the thirteen trace elements rose from \encImputeTraceRSqLinearObj{} to \encImputeTraceRSqLogObj{}, with no loss on the major elements. Table~\ref{tab:pretrain_hparams} shows the configuration and outcome of the pretraining run, while Fig.~\ref{fig:pretrain_loss} shows the loss curve.

\begin{table}[h]
\caption{Pretraining configuration and outcome.}
\label{tab:pretrain_hparams}
\centering
\begin{tabular}{lll}
\toprule
\multicolumn{1}{c}{\textbf{Hyperparameter}} &
\multicolumn{1}{c}{\textbf{Meaning}} &
\multicolumn{1}{c}{\textbf{Value}} \\
\midrule
\texttt{mask\_prob} & Fraction of populated rows masked & 0.281 \\
\texttt{batch\_size} & Rocks per step & 32 \\
\texttt{lr} & Adam learning rate & $8.18\times10^{-5}$ \\
\texttt{weight\_decay} & Adam weight decay & $1.98\times10^{-3}$ \\
\texttt{epochs} & Ceiling, not a target & 60 \\
\texttt{early\_stop\_patience} & Epochs without val improvement & 3 \\
\texttt{early\_stop\_min\_delta} & Improvement threshold & $1.0\times10^{-3}$ \\
\texttt{val\_split} / \texttt{test\_split} & Held-out fractions & 0.20 / 0.20 \\
\texttt{seed} & Run seed & 0 \\
Best epoch & When early stopping triggered & 16  \\
Train / val / test loss & Standardized log MSE & 0.199 / 0.200 / 0.193 \\
\bottomrule
\end{tabular}
\end{table}

\paragraph{Hyperparameter tuning.} Using the Optuna package \cite{akiba_optuna_2019}, two Bayesian searches were performed to optimize the encoder's hyperparameters. All trials were ranked by validation loss. Both searches were run on an earlier configuration that pretrained on only the igneous rocks of the compilation, and their selections were reused for the encoder trained on all rock types, as reported in Tables~\ref{tab:encoder_hparams} and~\ref{tab:pretrain_hparams}. The first search varied architecture and optimization jointly and selected the model width of 256, the 4 encoder layers, and the 8 attention heads reported in Table~\ref{tab:encoder_hparams}. The second search accompanied the adoption of the standardized log-space objective. It fixed the architecture and re-tuned the optimization values: the learning rate, mask probability, weight decay, dropout, and batch size. This search ran 30 trials of 20 epochs each, beginning from 10 random startup trials. The expansion of the pretraining set to all rock types and the addition of the redox row came after both searches and reused their selections without further tuning.

\subsection{Simulated label corpus}
\label{sec:corpus_appendix}

\paragraph{Composition selection.} Here, we use corpus to refer to the set of simulated labels that supervise the prediction heads. Each label pairs one measured whole-rock composition from the global geochemical compilation with a mineral assemblage calculated by \texttt{MAGEMin}. The latter is a Gibbs energy minimizer that returns the equilibrium assemblage of a composition at a specified temperature and pressure \cite{riel_magemin_2022}. We simulated igneous compositions, in-line with how the thermodynamic dataset implemented in \texttt{MAGEMin} covers igneous systems. 

The igneous compositions were narrowed in three passes: by rock name, by chemical range, and by the completeness of each analysis. By name, we kept compositions classified under volcanic rock names. We omitted the foidite family, for which the solver failed to converge across the sampled formation conditions. By chemical range, we excluded compositions with silica exceeding 80 wt\% or magnesia exceeding 18 wt\%. The silica bound was chosen to lie slightly beyond the 77 wt\% upper limit of the total alkali-silica classification \cite{bas_chemical_1986}. Meanwhile, the magnesia bound is the minimum content defining the most magnesium-rich volcanic rocks \cite{le_bas_iugs_2000}. Compositions beyond either bound fall outside what the corpus is meant to represent. For completeness, we required that a composition possess iron in at least one of its oxidation states, allowing the solver to resolve iron-bearing phases. We excluded compositions whose major-oxide sum fell below 90 wt\%, which may represent incomplete characterization.

\paragraph{Generator.} We called the \texttt{MAGEMin} solver, unmodified, through the Julia interface it distributes. Each composition was simulated under several draws, where a draw is one sampled pair of pressure and dissolved water content (Section \ref{sec:methods}). Every draw is evaluated along a cooling ladder, a fixed sequence of temperatures from 1200 to 600 $^\circ$C in 20 $^\circ$C steps at which the solver is called again for the same composition and pressure. A Python orchestrator samples the conditions of each draw, expands the cooling ladder into individual solver calls, and sends those calls to the solver in chunks. 

Each simulation follows one parcel of magma as it cools and crystallizes into mineral phases. At every temperature on the cooling ladder, the solver returns the assemblage that minimizes Gibbs free energy. We record the assemblage at fixed residual-melt fractions rather than at fixed temperatures, because compositions crystallize over different temperature ranges. In particular, we record each draw at 50 and 30 wt\% residual melt. Two values enable one cooling path to contribute assemblages at different extents of crystallization. We stop at 30 wt\% to remain away from the near-solidus region. 

For each recorded assemblage, we exclude the melt and any exsolved fluid from the mineral label before renormalizing the remaining solids to sum to one. The renormalized solids are the target of the assemblage head, the oxide composition of each solid is the target of the composition head, and the pressure, temperature, and dissolved water of the readout are the targets of the conditions head. We retain the melt and fluid fractions and their compositions, as the mass-balance term of the training objective reconstructs the input composition from every phase present (Appendix~\ref{sec:heads_appendix}). 

\paragraph{Iron speciation.} The solver requires the iron in each composition to be divided between its ferrous and ferric states. Where a composition reported both FeO and Fe$_2$O$_3$, we used the measured ratio. When only total iron was reported, we assigned a ferric fraction by rock type following two published compilations \cite{le_maitre_problems_1976, middlemost_iron_1989}. The fractions range from 0.15 for basalt to 0.45 for the most silica-rich classes, within the bounds the two compilations recommend for each class. Fixing the iron ratio in turn fixed the oxygen content of each simulation, so we did not sample oxidation state as a third condition alongside pressure and dissolved water.

\paragraph{Broad condition sampling.} We sampled the pressure and water content for every draw from an intentionally broad distribution:

\begin{equation}\label{eq:proposal}
P \sim \mathrm{Uniform}[0.5,\, 12]\ \mathrm{kbar},
\qquad
X_{\mathrm{H_2O}} \sim \mathrm{Uniform}\big[0,\, \min(8,\, S(P))\big]\ \mathrm{wt\%}
\end{equation}

where $S(P)$ is the pure-H$_2$O saturation curve for a rhyolitic melt implemented in VolatileCalc \cite{newman_volatilecalc_2002}, which we evaluated at 850 $^\circ$C and applied to all compositions. We further capped the draw at 8 wt\%, the upper end of the water contents recorded in melt inclusions from arc magmas \cite{wallace_volatiles_2005, plank_why_2013}. Otherwise, the solubility curve would assign far more water at the upper end of our pressure range. While we formulated the foregoing treatment for ease of corpus generation, we acknowledge a gap for alkaline compositions. Specifically, phonolitic melts dissolve more water than rhyolitic melts at the same pressure \cite{carroll_solubility_1997}. As a result, wet, low-pressure conditions are undersampled for those alkaline compositions. Opportunities for future improvements to the methodology include implementing a composition-aware solubility model.

\paragraph{Phase vocabulary.} The solver outputs phases according to the solution models in its thermodynamic dataset. We performed three small adjustments to create prediction targets from those phases, where the net vocabulary came out to $J = \sysPhaseVocab{}$ phases. Firstly, tridymite, cristobalite, and quartz are polymorphs, so we collapsed all three into one mineral. We note that tridymite and cristobalite only appear in roughly 2 of every 10{,}000 draws. But when they are present, they carry about half the crystal mass present. Meanwhile, we kept the aluminosilicate polymorphs separate because their identity reflects the pressure at which they formed, which the conditions head predicts from the phases present. Table~\ref{tab:phase_vocab} lists the resulting vocabulary.

\begin{table}[h]
\caption{The \sysPhaseVocab{}-phase vocabulary of the assemblage head, with each phase's presence rate, defined as the percent of training labels in which the phase is present (i.e., holds more than 0.1 wt\% of the crystal mass). The ten phases above the divider map onto the \rrBenchGroups{} benchmark categories of Table~\ref{tab:magemin_collapse}.}
\label{tab:phase_vocab}
\centering
\small
\begin{tabular}{llcr}
\toprule
\multicolumn{1}{c}{\textbf{Phase}} &
\multicolumn{1}{c}{\textbf{Abbreviation}} &
\multicolumn{1}{c}{\textbf{Benchmark category}} &
\multicolumn{1}{c}{\textbf{Presence rate (\%)}} \\
\midrule
plagioclase     & plag & Feldspar     & \pvPlag \\
alkali feldspar & afs  & Feldspar     & \pvAfs \\
clinopyroxene   & cpx  & Pyroxene     & \pvCpx \\
orthopyroxene   & opx  & Pyroxene     & \pvOpx \\
amphibole       & amp  & Amphibole    & \pvAmp \\
biotite         & bi   & Biotite      & \pvBi \\
olivine         & ol   & Olivine      & \pvOl \\
spinel          & spl  & Fe-Ti oxides & \pvSpl \\
ilmenite        & ilm  & Fe-Ti oxides & \pvIlm \\
quartz          & q    & Quartz       & \pvQ \\
\midrule
rutile          & ru   & --           & \pvRu \\
garnet          & g    & --           & \pvG \\
titanite        & sph  & --           & \pvSph \\
muscovite       & mu   & --           & \pvMu \\
epidote         & ep   & --           & \pvEp \\
cordierite      & cd   & --           & \pvCd \\
sillimanite     & sill & --           & \pvSill \\
nepheline       & ne   & --           & \pvNe \\
kyanite         & ky   & --           & \pvKy \\
andalusite      & and  & --           & \pvAnd \\
corundum        & cor  & --           & \pvCor \\
\bottomrule
\end{tabular}
\end{table}

Secondly, chlorite and ferropericlase appear in roughly 1 to 2 of every 10{,}000 draws, carrying negligible mass when they do. A phase this rare offers the assemblage head too few examples to learn from, so both were dropped. Lastly, the solver returns feldspar as a single solution phase, although plagioclase and alkali feldspar coexist as separate crystals in 13 to 19\% of feldspar-bearing draws. Averaging two coexisting minerals into one entry produces an aphysical composition that would occur in real rocks. To avoid the composition head training on this single solution phase feldspar, we split each lumped instance into plagioclase or alkali feldspar according to its composition. We compute the molar fractions of the calcium and potassium endmembers and compare them against two curves in the anorthite-orthoclase plane that delineate plagioclase and alkali feldspar \cite{shi_mineralml_2026}. A lumped instance inside either field is assigned to that mineral, and a lumped instance falling between the two fields is assigned to whichever of the two components is larger.

\paragraph{Label construction.} \texttt{MAGEMin} reports phases at arbitrarily small abundances, requiring us to define a threshold to determine when a phase should count as present or absent. Specifically, we defined the threshold at 0.1 wt\%, roughly an order of magnitude below the detection limits of analytical techniques such as X-ray diffraction and optical point counting. Each label is therefore a binary indicator vector over the $J = \sysPhaseVocab{}$ vocabulary phases, with a 1 denoting when a phase is present. The labels show a median of 5 phases and a 10th-to-90th percentile range of 3 to 7 phases. 

\paragraph{Scale and splits.} Table~\ref{tab:corpus_scale} shows the size of each split of the corpus. The realized counts are slightly below the target counts because a small number of compositions fail to converge. We split the corpus by composition rather than by row, and stratified by QAPF rock name. Rows generated from one composition share the same input chemistry, differing only in their sampled conditions and melt fractions. Thus, a row-level split would allocate the same rock to both training and evaluation. A floor of 30 compositions per class in the training split, and 5 in the validation and test splits, keeps rare classes represented. The \sysCorpusValBulks{} validation compositions constitute the held-out simulated set reported throughout Section~\ref{sec:results}.

\begin{table}[h]
\caption{Size of the simulated label corpus by split.}
\label{tab:corpus_scale}
\centering
\begin{tabular}{lllll}
\toprule
\multicolumn{1}{c}{\textbf{Split}} &
\multicolumn{1}{c}{\textbf{Compositions}} &
\multicolumn{1}{c}{\textbf{Draws each}} &
\multicolumn{1}{c}{\textbf{Melt fractions read (wt\%)}} &
\multicolumn{1}{c}{\textbf{Labels}} \\
\midrule
Train & \sysCorpusTrainBulksRealized{} & \sysCorpusTrainDraws{} & 50, 30 & \sysCorpusTrainRows{} \\
Validation & \sysCorpusValBulks{} & \sysCorpusValtestDraws{} & 50, 30 & \sysCorpusValRows{} \\
Test & \sysCorpusTestBulks{} & \sysCorpusValtestDraws{} & 50, 30 & \sysCorpusTestRows{} \\
\bottomrule
\end{tabular}
\end{table}

\subsection{Prediction heads}
\label{sec:heads_appendix}

Three heads read the frozen encoder's pooled latent representation $h$ and produce the model's predictive distributions. Each is a small multilayer perceptron composed of a trunk of linear layers with GELU activations and dropout, followed by linear output maps. Table~\ref{tab:head_arch} summarizes each head, with parameter counts reflecting the trained model evaluated throughout this work.

\begin{table}[h]
\caption{The three prediction heads, where trunk width and depth were tuned by Bayesian search alongside the loss weights and optimizer settings (Appendix~\ref{sec:training_appendix}).}
\label{tab:head_arch}
\centering
\small
\begin{tabular}{lllll}
\toprule
\multicolumn{1}{c}{\textbf{Head}} &
\multicolumn{1}{c}{\textbf{Predicts}} &
\multicolumn{1}{c}{\textbf{Distribution}} &
\multicolumn{1}{c}{\textbf{Trunk}} &
\multicolumn{1}{c}{\textbf{Parameters}} \\
\midrule
Assemblage & phase presence and abundance & Bernoulli $\times$ Dirichlet & $\hdModeLayers \times \hdModeHidden$ & \sysModeHeadParams{} \\
Composition & per-phase oxide composition & Dirichlet per phase & $\hdCompLayers \times \hdCompHidden$ & \sysCompHeadParams{} \\
Conditions & $T$, $P$, dissolved H$_2$O & Gaussian mixture per axis & $\hdCondLayers \times \hdCondHidden$ & \sysCondHeadParams{} \\
\bottomrule
\end{tabular}
\end{table}

\paragraph{Assemblage head.} The two factors of Eq.~\ref{eq:hurdle} are separate branches trained on separate targets. We factorized the distribution because a single Dirichlet over the full \sysPhaseVocab-phase simplex cannot represent sparse labels. Specifically, a single Dirichlet cannot express an absent phase, as it must assign every phase a positive abundance. Factorization allows the presence branch to decide which phases are absent, enabling the Dirichlet layer to only describe present phases. The presence branch reads the latent representation $h$ alone and yields one logit per phase. The abundance branch meanwhile reads $h$ concatenated with a support vector $s$, a binary indicator of the present phases, and yields one concentration per phase through a softplus. At training, $s$ is the label's true support, while at inference, $s$ is sampled from the presence branch.

\paragraph{Composition head.} For every phase $j$, the head models the oxide composition $x_j$ as a Dirichlet over the \sysCompOxides{} oxides reported by the simulator (in weight percent):

\begin{equation}\label{eq:comp_head}
    p(x_j \mid h) = \mathrm{Dir}\big(x_j \,;\, c_j\,\mu_j\big),
    \qquad
    \mu_j = \mathrm{softmax}(z_j),
    \qquad
    c_j = \mathrm{softplus}(\tau_j)
\end{equation}

where the head's trunk maps the latent representation $h$ to the outputs $z_j$ and $\tau_j$. The mean $\mu_j$ is the vector of predicted oxide proportions and the concentration $c_j$ describes the width of the distribution. The loss is evaluated only on the phases present in each label, mirroring the abundance factor. Because a Dirichlet cannot express an exact zero, oxides reported as zero by the simulator are clipped to  \hdCompEps{} and the target renormalized.

\paragraph{Conditions head.} For each formation condition (temperature, pressure, and dissolved water content), the head predicts an independent \hdCondComponents-component Gaussian mixture:

\begin{equation}\label{eq:cond_head}
    p(y_a \mid h, s) = \sum_{k=1}^{\hdCondComponents} \pi_{a,k}\,
    \mathcal{N}\big(y_a \,;\, \mu_{a,k},\, \sigma_{a,k}^2\big),
    \qquad
    a \in \{T,\ P,\ X_{\mathrm{H_2O}}\}
\end{equation}

where $a$ indexes the formation condition and $y_a$ its value. The head's trunk produces three linear outputs per condition:  the means $\mu_a$, and pre-activations $u_a$ and $v_a$ for the weights ($\pi_a$) and standard deviations ($\sigma_a$), respectively. 

\begin{equation}\label{eq:cond_links}
    \pi_a = \mathrm{softmax}(u_a),
    \qquad
    \sigma_a = \mathrm{softplus}(v_a)
\end{equation}

As with the assemblage head, the conditions head reads the latent representation $h$ concatenated with the support $s$, where the support is teacher-forced at training and sampled at inference. The conditions head predicts each condition in its raw physical units (°C, kbar, and wt\%). To match the scale of the data, we initialize the biases of $\mu_a$ and $v_a$ from each condition's target mean and standard deviation.

\paragraph{Mass balance.} An additional loss term constrains the heads jointly by reconstructing the input composition $B$ from their predictions. A corpus label describes a partly crystallized magma, where the predicted solids coexist with the remaining melt and any exsolved fluid. The model predicts the solid portion while the melt and fluid come from the label directly. The reconstruction $\hat{B}$ weights each portion by its mass fraction and sums them: 

\begin{equation}\label{eq:closure}
  \hat{B} = \big(1 - F_{\mathrm{liq}} - F_{\mathrm{fl}}\big) \sum_{j} m_j\, C_j
  \;+\; F_{\mathrm{liq}}\, C_{\mathrm{liq}} + F_{\mathrm{fl}}\, C_{\mathrm{fl}}
\end{equation}

where $m_j$ is the assemblage head's mean abundance for phase $j$ on the true support and $C_j$ is the composition head's mean composition. The mass fractions $F_{\mathrm{liq}}$ and $F_{\mathrm{fl}}$ give the melt and exsolved fluid, and $C_{\mathrm{liq}}$ and $C_{\mathrm{fl}}$ their compositions. Because the melt and fluid quantities come from the label, any mismatch between $\hat{B}$ and $B$ must be reduced by adjusting the predicted abundances and compositions.

\subsection{Training}
\label{sec:training_appendix}

The three heads train jointly on the frozen encoder, where the objective combines five terms: an NLL for each predictive factor of Appendix~\ref{sec:heads_appendix}, and the mass-balance penalty $\mathcal{L}_{\mathrm{balance}}$:

\begin{equation}\label{eq:joint_loss}
\mathcal{L} = \mathcal{L}_{\mathrm{presence}}
+ \lambda_{\mathrm{abund}}\, \mathcal{L}_{\mathrm{abundance}}
+ \lambda_{\mathrm{comp}}\, \mathcal{L}_{\mathrm{comp}}
+ \lambda_{\mathrm{cond}}\, \mathcal{L}_{\mathrm{cond}}
+ \lambda_{\mathrm{balance}}\, \mathcal{L}_{\mathrm{balance}}
\end{equation}

Describing each of the loss terms in order, we have $\mathcal{L}_{\mathrm{presence}}$, the binary cross-entropy between the Bernoulli probabilities $p_j$ of Eq.~\ref{eq:hurdle} and the label's support. $\mathcal{L}_{\mathrm{abundance}}$ is the NLL of the Dirichlet in Eq.~\ref{eq:hurdle}, evaluated under the label's true support. $\mathcal{L}_{\mathrm{comp}}$ is the NLL of Eq.~\ref{eq:comp_head}, scored on the present phases alone. $\mathcal{L}_{\mathrm{cond}}$ is the NLL of the mixtures in Eq.~\ref{eq:cond_head}, summed over temperature, pressure, and water content. And lastly, $\mathcal{L}_{\mathrm{balance}}$ is the squared residual of Eq.~\ref{eq:closure}.

\paragraph{Hyperparameter search.} A single Bayesian search in Optuna \cite{akiba_optuna_2019} tuned seven variables jointly: the learning rate, the weight decay, $\lambda_{\mathrm{abund}}$, $\lambda_{\mathrm{balance}}$, the trunk dropout of all heads, and the assemblage trunk's width and depth (Table~\ref{tab:heads_hpo}). The search ran \hpoTrials{} trials at a \hpoTrialEpochs-epoch ceiling, beginning from \hpoStartupTrials{} random startup trials, and ranked each trial on the assemblage head's raw, unweighted validation term at its own best epoch. Differences of a few hundredths of a nat fall within the validation evaluation's sampling noise, so trials within \hpoTieBand{} nats of the leader count as tied. To settle a tie among the leading trials, we re-evaluate each tied trial on supports sampled from its own presence branch (in contrast to the teacher-forced objective that ranked the trials). The winning trial was retrained at the full \hdEpochsCeiling-epoch ceiling to yield the checkpoint used throughout this work.

Because $\mathcal{L}_{\mathrm{presence}}$ averages over the \sysPhaseVocab{} phases while $\mathcal{L}_{\mathrm{abundance}}$ sums over the support, the Optuna-selected $\lambda_{\mathrm{abund}} = \hdMagWeight{}$ corresponds to an effective weight of $21 \times \lambda_{\mathrm{abund}} = \hdEffLambda{}$. Similarly, $\lambda_{\mathrm{cond}} = \hdLambdaCond{}$ primarily cancels the raw °C scale of the temperature term rather than down-weighting the conditions head. Optimization uses AdamW at learning rate \hdLr{}, with weight decay \hdWeightDecay{}, batch size \hdBatchSize{}, and seed \hdSeed{}. Gradients are clipped per head rather than under one shared norm threshold.

We train the heads on the \sysCorpusTrainRows{} training-split rows of Table~\ref{tab:corpus_scale} and select hyperparameters on the \sysCorpusValRows{} rows of the validation compositions. We save each head at the epoch where its raw, unweighted validation term is best. Training stops once every head is \hdPatience{} epochs past its own best. The final training run stopped at epoch \hdEpochsRun{}, saving the assemblage head at epoch \hdModeBestEpoch{}, the composition head at \hdCompBestEpoch{}, and the conditions head at \hdCondBestEpoch{} (Fig.~\ref{fig:heads_loss}).

\begin{table}[h]
\caption{Joint search space and optimized values over the training environment and assemblage-head architecture. The remaining weights were fixed during the search at $\lambda_{\mathrm{comp}} = \hdLambdaComp{}$ and $\lambda_{\mathrm{cond}} = \hdLambdaCond{}$.}
\label{tab:heads_hpo}
\centering
\small
\begin{tabular}{llll}
\toprule
\multicolumn{1}{c}{\textbf{Hyperparameter}} &
\multicolumn{1}{c}{\textbf{Meaning}} &
\multicolumn{1}{c}{\textbf{Range}} &
\multicolumn{1}{c}{\textbf{Selected}} \\
\midrule
\texttt{lr} & AdamW learning rate & \hpoRangeLr{} & \hdLr{} \\
\texttt{weight\_decay} & AdamW weight decay & \hpoRangeWeightDecay{} & \hdWeightDecay{} \\
$\lambda_{\mathrm{abund}}$ & Abundance weight, nominal & \hpoRangeMagWeight{} & \hdMagWeight{} \\
$\lambda_{\mathrm{balance}}$ & Mass-balance weight & \hpoRangeLambdaBalance{} & \hdLambdaBalance{} \\
\texttt{dropout} & Trunk dropout, all heads & \hpoRangeDropout{} & \hdDropout{} \\
\texttt{mode\_hidden} & Assemblage trunk width & \hpoRangeModeHidden{} & \hdModeHidden{} \\
\texttt{mode\_layers} & Assemblage trunk depth & \hpoRangeModeLayers{} & \hdModeLayers{} \\
\bottomrule
\end{tabular}
\end{table}

\subsection{Recalibration layer}
\label{sec:recalibrator_appendix}

The recalibration layer is a fitted map on the assemblage head's two outputs: the presence logits and the Dirichlet concentrations, parameterized over the $G = \sysRealGroups{}$ mineral categories in the benchmark set. The vectors $(a, b, c)$ of Eq.~\ref{eq:recal} are length $G$ and its matrices $(A, B)$ are $G \times G$. Two further parameters add a silica slope to the presence logits of quartz and amphibole.

In the layer used throughout this work, the abundance factor is a two-component Dirichlet mixture whose components share a mean: 

\begin{equation}\label{eq:contam}  
    p(m) = (1 - \epsilon)\,\mathrm{Dir}(m \mid \alpha') + \epsilon\,\mathrm{Dir}(m \mid k\,\alpha')                              
\end{equation} 

Scaling every concentration by $k$ preserves the mean $\alpha' / \sum_g \alpha'_g$. With the fitted $k < 1$, the total falls from $\sum_g \alpha'_g$ to $k \sum_g \alpha'_g$, thereby widening $\mathrm{Dir}(m \mid k\,\alpha')$ about that mean. With the maps of Eq.~\ref{eq:recal} fitted, the predictions are centered on the measured abundances but too confident, in that the measured abundance of a minority of rocks lies outside the predicted interval. A single Dirichlet has one width and could accommodate those rocks only by widening every prediction, whereas the mixture confines the widening to a component of weight $\epsilon$, leaving the remaining $1 - \epsilon$ of the predicted distribution unchanged.

We define three configurations of increasing capacity, each containing the previous: a per-mineral category tier of \sysRecalibVectorParams{} parameters, a mineral category-coupled tier of \sysRecalibMatrixParams{} that adds $A$ and $B$, and a mixture tier containing the mixture of Dirichlet distributions. Each initializes to the identity map, with $a$ at one, with $b$, $c$, $A$, $B$, and the silica slopes at zero, and with $k = 1$, which collapses the mixture to a single Dirichlet. An unfit layer therefore reproduces the zero-shot prediction.

The layer is fitted by minimizing the NLL of Eq.~\ref{eq:hurdle}. Optimization uses full-batch Adam at learning rate \rcFitLr{} for \sysRecalibEpochs{} steps with no early stopping.
Every parameter of the layer is fitted, including the mixture tier's $\epsilon$ and $k$, which reach \rrTailedEps{} and \rrTailedK{}, respectively. Weight decay is evaluated on the validation set once for each configuration, over a grid from \rcWdGridLo{} to \rcWdGridHi{}. At each candidate value, the layer is fitted on 25, 50, 100, and 200 rocks, three replicates per size, and evaluated on the validation set. The candidate yielding the lowest mean score is used for every subsequent fit of that configuration. The mixture tier, with \sysRecalibParams{} parameters, is used throughout this work, with Appendix~\ref{sec:resupervision_appendix} comparing the performance of the three tiers.

\subsection{Measured-rock benchmark construction}
\label{sec:benchmark_appendix}

The USGS surveys \cite{du_bray_petrographic_2016, du_bray_petrographic_2019} grade each rock on a five-point alteration scale.  To construct the benchmark, we reasoned that grade 1 and 2 rocks would reflect their parent magma without being pristine, while grade 3 and above would reflect hydrothermal replacement of the parent assemblage that no thermodynamic simulation could capture. Thus, the benchmark comprises the \rrBenchN{} rocks graded 1 or 2, while rocks graded 3--5 are excluded.

\begin{table}[h]
\caption{Mapping predicted phases onto the \rrBenchGroups{} mineral
categories quantified in \cite{du_bray_petrographic_2016} and \cite{du_bray_petrographic_2019}.}
\label{tab:magemin_collapse}
\centering
\small
\begin{tabular}{ccccccc}
\toprule
\multicolumn{1}{c}{\textbf{Feldspar}} &
\multicolumn{1}{c}{\textbf{Pyroxene}} &
\multicolumn{1}{c}{\textbf{Amphibole}} &
\multicolumn{1}{c}{\textbf{Biotite}} &
\multicolumn{1}{c}{\textbf{Olivine}} &
\multicolumn{1}{c}{\textbf{Fe-Ti oxides}} &
\multicolumn{1}{c}{\textbf{Quartz}} \\
fsp & px & amp & bi & ol & opq & q \\
\midrule
plagioclase     & clinopyroxene & amphibole & biotite & olivine & spinel   & quartz \\
alkali feldspar & orthopyroxene &           &         &         & ilmenite &        \\
\bottomrule
\end{tabular}
\end{table}

We split the benchmark by stratigraphic unit into train, validation, and test sets, where each stratigraphic unit is assigned whole. We chose stratigraphic units as the basis for data splits given that rocks within a stratigraphic unit possess similar chemistry. Thus, splitting by unit prevents evaluating rocks from groups used for training. The target size of each set is 60\% (train), 15\% (val), and 25\% (test) of the \rrBenchN{} rocks. Stratigraphic units are sorted by the number of rocks they contain and assigned to train/val/test sets, where the largest unassigned unit goes to the set furthest below its target. While the procedure is deterministic, ties arise when sets are the same number of rocks away from their target. We generate 400 candidate splits by rerunning the procedure with ties broken at random. To filter the candidate sets down to one, we reasoned that silica content and mineral occurrence governed the model's input and labels, respectively. The former determines volcanic rock compositions (e.g., mafic, felsic) while the latter determines the occurrence rate that presence predictions are compared to. We selected the split possessing train/val/test sets that most closely matched the full benchmark with respect to mineral occurrence and silica distribution as follows: 

\begin{equation}
    D = \max_{s,g} \left| r_{s,g} - r_g \right| + \max_{s,p} \frac{\left| q_{s,p} - q_p \right|}{R}
    \label{eq:split_balance}
\end{equation}

where $D$ is the deviation score minimized over the 400 candidates, $s$ indexes the train, validation, and test sets, and $g$ indexes the \rrBenchGroups{} mineral categories. The term $r_{s,g}$ is the occurrence rate of mineral category $g$ within set $s$ and $r_g$ is its rate across the full benchmark. Likewise $q_{s,p}$ is the $p$-th SiO$_2$ percentile of set $s$ and $q_p$ is that percentile across the full benchmark, taken at $p \in \{5, 50, 95\}$. Lastly, $R$ is the SiO$_2$ range of the full benchmark. The resulting split contains \rrTrainNRocks{}, \rrValNRocks{}, and \rrTestNRocks{} rocks over \rrTrainNUnits{}, \rrValNUnits{}, and \rrTestNUnits{} stratigraphic units.

The sources report modal abundance as volume percent, while the assemblage head predicts mass fractions (reflecting the units used in \texttt{MAGEMin}). Each reported mineral therefore converts through a representative density (Table~\ref{tab:mineral_density}) before collapsing onto the mineral categories of Table~\ref{tab:magemin_collapse}. Densities are obtained from The Mineralogy Database \cite{barthelmy_mineralogy_2012}, where a label spanning several minerals (e.g., undivided feldspar or the Fe-Ti oxides) is assigned an intermediate value among its members. For a solid solution, the density is interpolated between end members at the expected composition, e.g., olivine at Fo$_{80}$ is approximated as $\rho = 0.8\,\rho_{\mathrm{Fo}} + 0.2\,\rho_{\mathrm{Fa}} = 3.50$. Labels with no composition to interpolate along are assigned the mean of their members. When the surveys mark a mineral as trace (`TR' or $<$1), we count the mineral as present, but without a numeric abundance. The converted masses then re-normalize to abundances over the numerically reported categories. For amphibole, the fitted abundance correction is small enough that its sign flips within the plausible density range. A fitted value would therefore reflect the density assumption rather than a model error. As a result, we fix the amphibole correction at zero in the recalibration layer (Appendix~\ref{sec:recalibrator_appendix}).

On the benchmark, we expand each measured assemblage into a support over the \sysPhaseVocab{} phases that the abundance factor requires. For a mineral category absent from the rock, every member phase is switched off. Within a mineral category that is present, we switch on the member phases whose predicted presence probability exceeds 0.5. When no member exceeds 0.5, only the single most likely member is switched on. For the eleven phases outside the categories, we take the presence predictions as the support.

\begin{table}[H]
\centering
\caption{Densities used to convert reported volume percent to mass fractions.}
\label{tab:mineral_density}
\begin{tabular}{llr}
\toprule
\multicolumn{1}{c}{\textbf{Reported mineral}} &
\multicolumn{1}{c}{\textbf{Category}} &
\multicolumn{1}{c}{\textbf{Density (g\,cm$^{-3}$)}} \\
\midrule
plagioclase & feldspar & 2.68 \\
alkali feldspar & feldspar & 2.57 \\
feldspar & feldspar & 2.63 \\
quartz & quartz & 2.65 \\
silica phases & quartz & 2.32 \\
pyroxene & pyroxene & 3.40 \\
augite & pyroxene & 3.40 \\
hypersthene & pyroxene & 3.45 \\
bronzite & pyroxene & 3.35 \\
hornblende & amphibole & 3.15 \\
cummingtonite & amphibole & 3.30 \\
biotite & biotite & 3.05 \\
olivine & olivine & 3.50 \\
Fe-Ti oxides & Fe-Ti oxides & 4.94 \\
\bottomrule
\end{tabular}
\end{table}

\subsection{Presence discrimination}
\label{sec:discrimination_appendix}

A mineral category is marked as present when the sources report any of its member minerals, including those marked as trace. A reported abundance of zero is counted as absent. The predicted probability is the model's presence probability for the category, collapsed from the \sysPhaseVocab{} phases onto the \rrBenchGroups{} mineral categories. Area under the receiver operating characteristic curve (AUROC) describes the probability that a rock containing a mineral category receives a higher predicted probability than a rock lacking it. A value of 0.5 marks a random classifier, while 1.0 reflects perfect discrimination of present from absent rocks.

The whiskers in Fig.~\ref{fig:cohort_card}b are obtained by bootstrapping the benchmark set of rocks while holding the model fixed. Each bootstrapped set is created by sampling stratigraphic units with replacement. In particular, the stratigraphic units are sampled \rrBenchNUnits{} times, matching the number of units in the benchmark set. As with the data splits (Appendix~\ref{sec:benchmark_appendix}), units are sampled whole rather than individual rocks, since treating rocks of similar chemistry as independent observations would understate the spread. A total of 1000 bootstrapped sets are created through the foregoing procedure, with each set having its own AUROC values. Each whisker reported in Fig.~\ref{fig:cohort_card}b spans the 25th to 75th percentile of the 1000 AUROC values.


\newpage
\section{Composition and conditions predictions}
\label{sec:compcond_appendix}
\renewcommand{\thefigure}{C\arabic{figure}}
\renewcommand{\theHfigure}{C\arabic{figure}}
\setcounter{figure}{0}

\begin{figure}[h]
    \centering
    \includegraphics[width=0.85\linewidth]{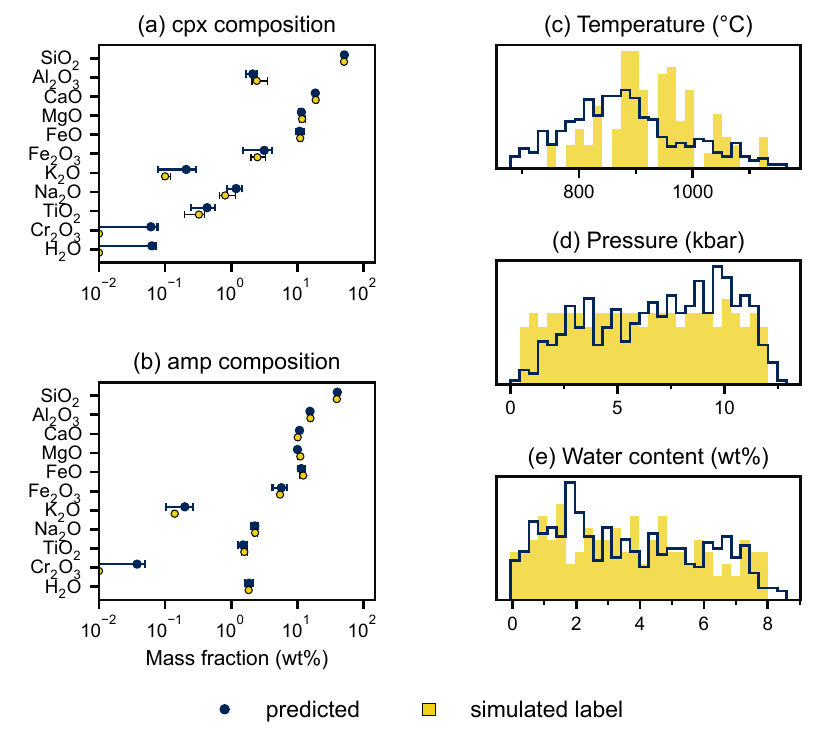}
    \caption{\textbf{Composition and conditions predictions against simulated labels.} Predictions for one validation composition, an andesite, was selected because it had the median of the per-composition assemblage NLL at the 50\%-melt readout. \textbf{(a, b)} Predicted compositions of the most probable anhydrous phase (clinopyroxene, a) and the most probable hydrous phase (amphibole, b), as ranked by the presence branch. Predicted points represent the mean over \acCompDraws{} candidate compositions from the predicted Dirichlet and whiskers span the 25th to 75th percentile. Simulated label points reflect the median of the simulated labels across the composition's \sysCorpusValtestDraws{} draws and whiskers span the 25th to 75th percentile. \textbf{(c--e)} Predicted formation temperature, pressure, and dissolved water content over \acCondSamples{} assemblages sampled from the presence branch, against the sampled conditions of the same \sysCorpusValtestDraws{} draws.}
    \label{fig:compcond_sim}
\end{figure}

\begin{figure}[h]
    \centering
    \includegraphics[width=0.9\linewidth]{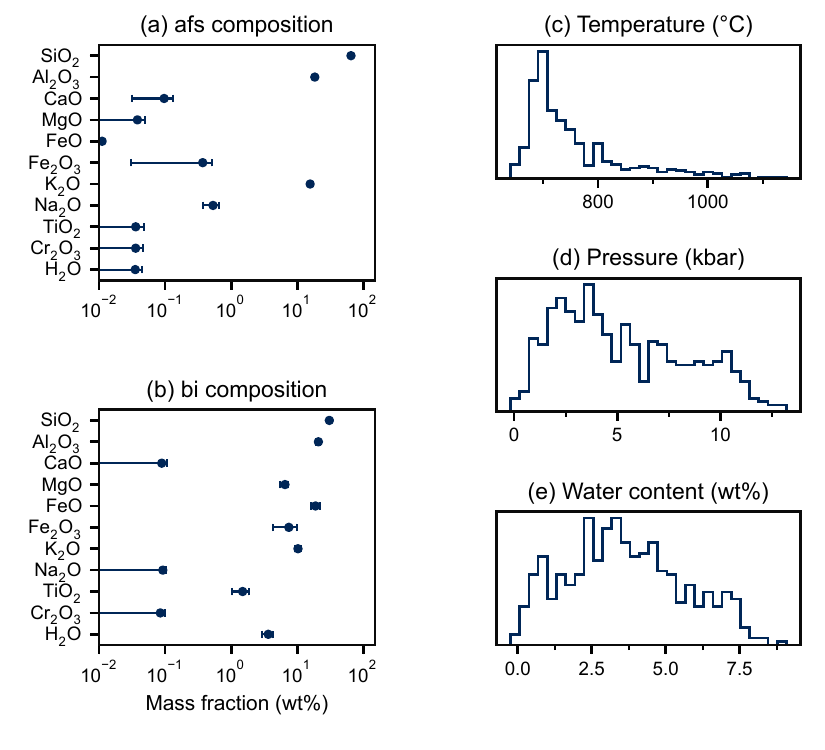}
    \caption{\textbf{Zero-shot composition and conditions predictions for a measured rock.} Predictions for one benchmark rhyolite, the rock of Fig.~\ref{fig:inference_card}, selected for having the median of the recalibrated joint NLL over the test set. \textbf{(a, b)} Predicted composition of the most probable anhydrous phase (alkali feldspar, panel a) and the most probable hydrous phase (biotite, panel b), as ranked by the presence branch. Points represent the mean over \acCompDraws{} candidate compositions from the predicted Dirichlet and whiskers span the 25th to 75th percentile. \textbf{(c--e)} Predicted formation temperature, pressure, and dissolved water content over \acCondSamples{} assemblages sampled from the presence branch. The benchmark reports neither mineral compositions nor formation conditions, so no measured counterpart appears.}
    \label{fig:compcond_real}
\end{figure}

Section~\ref{sec:results} evaluates the assemblage head, but not the composition or conditions head, because the benchmark contains modal mineralogy but not oxide compositions or formation conditions. The composition and conditions head nonetheless produce predictions for every rock. Fig.~\ref{fig:compcond_sim} begins by illustrating the predicted oxide compositions and formation-condition distributions on one held-out validation composition, where the simulated labels supply a reference. The composition, an andesite, was selected because it had the median of the per-composition assemblage NLL at the 50\%-melt readout. The predicted compositions agree with the labels for anhydrous (Fig.~\ref{fig:compcond_sim}a) and hydrous phases alike. For example, Fig.~\ref{fig:compcond_sim}b shows that the structural water that amphibole carries in its lattice -- distinct from the melt's dissolved water content predicted by the conditions head -- is recovered by the composition head. In addition, the predicted temperature distribution (Fig.~\ref{fig:compcond_sim}c) follows the labels, while the predicted pressure and water distributions (Fig.~\ref{fig:compcond_sim}d-e) reproduce the breadth of the sampled conditions described by Eq.~\ref{eq:proposal}. We note that the width of these posteriors is set by our choice of condition sampling rather than by the rock itself, which acts as a source of bias to the model (e.g., Appendix~\ref{sec:resupervision_appendix} traces the confidence gap on real rocks to condition sampling). Lastly, Fig.~\ref{fig:compcond_real} shows the same predictions for the rhyolite of Fig.~\ref{fig:inference_card}. While no labels exist for the predictions, we observe that the model predicts formation temperatures concentrated at the cool end of the sampled range, consistent with the lower crystallization temperatures of silica-rich magmas.

\clearpage
\section{Comparing simulated and measured labels}
\label{sec:resupervision_appendix}

\renewcommand{\thefigure}{D\arabic{figure}}
\renewcommand{\theHfigure}{D\arabic{figure}}
\setcounter{figure}{0}

\begin{figure}[H]
    \centering
    \includegraphics[width=0.95\linewidth]{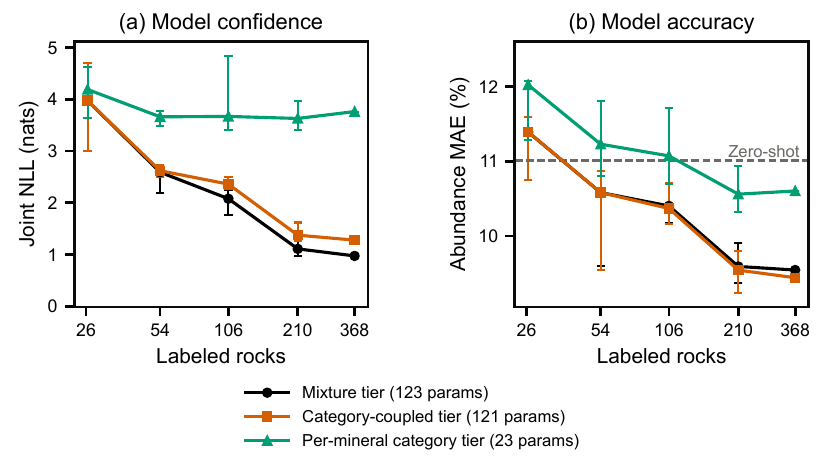}
    \caption{\textbf{Recalibration tiers across fit sizes.} The three recalibration configurations described in Appendix~\ref{sec:recalibrator_appendix}. \textbf{(a)} Median joint NLL. For reference, the zero-shot joint NLL is \rrValZsProd{} nats, above the plotted range. \textbf{(b)} Abundance MAE, reported as the mean of the \rrBenchGroups{} per-category MAE values, where the dashed line marks the zero-shot value. For both panels, error bars span the interquartile range over the \fsBudgetDraws{} replicates at each fitting size, evaluated on the fixed validation set of \rrValNRocks{} rocks spanning \rrValNUnits{} stratigraphic units. Axes are shared with Figs.~\ref{fig:fewshot_heads} and~\ref{fig:fewshot_strongest}.}
    \label{fig:fewshot_tiers}
\end{figure}

\begin{figure}[h]
    \centering
    \includegraphics[width=0.95\linewidth]{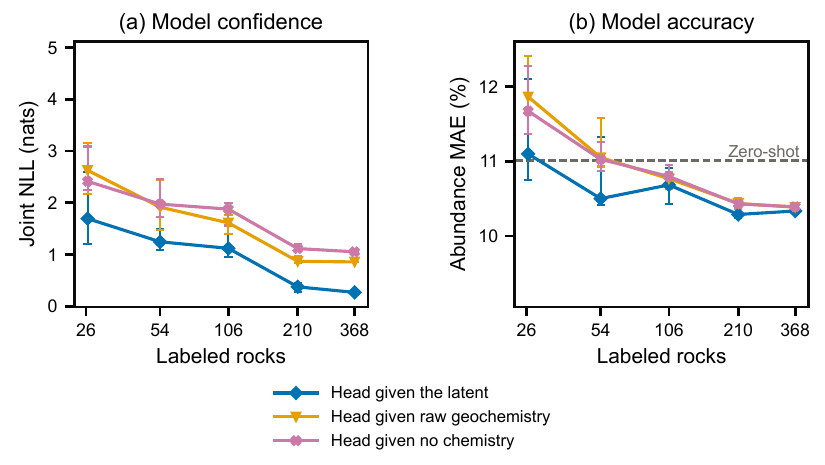}
    \caption{\textbf{Heads trained on measured abundances across fit sizes.} The three heads differing only in their input: the frozen encoder's latent representation, the composition the latent representation was computed from, or a zero vector. \textbf{(a)} Median joint NLL. For reference, the zero-shot joint NLL is \rrValZsProd{} nats, above the plotted range. \textbf{(b)} Abundance MAE, reported as the mean of the \rrBenchGroups{} per-category MAE values, where the dashed line marks the zero-shot value. For both panels, error bars span the interquartile range over the \fsBudgetDraws{} replicates at each fitting size, evaluated on the fixed validation set of \rrValNRocks{} rocks spanning \rrValNUnits{} stratigraphic units. Axes are shared with Figs.~\ref{fig:fewshot_tiers} and~\ref{fig:fewshot_strongest}.}
    \label{fig:fewshot_heads}
\end{figure}

\begin{figure}[H]
    \centering
    \includegraphics[width=1.0\linewidth]{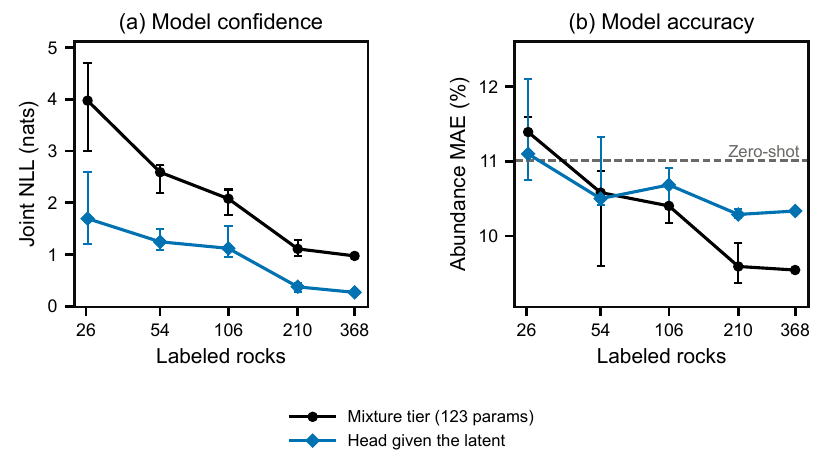}
    \caption{\textbf{Simulated and measured supervision across fit sizes.} The recalibration configuration is the mixture tier described in Appendix~\ref{sec:recalibrator_appendix}, which corrects predictions supervised by simulated labels. Meanwhile the head is the one given the latent representation, which is trained on measured abundances alone. Both represent the most performant from their respective groups in Figs.~\ref{fig:fewshot_tiers} and~\ref{fig:fewshot_heads}. \textbf{(a)} Median joint NLL. For reference, the zero-shot joint NLL is \rrValZsProd{} nats, above the plotted range. \textbf{(b)} Abundance MAE, reported as the mean of the \rrBenchGroups{} per-category MAE values, where the dashed line marks the zero-shot value. For both panels, error bars span the interquartile range over the \fsBudgetDraws{} replicates at each fitting size, evaluated on the fixed validation set of \rrValNRocks{} rocks spanning \rrValNUnits{} stratigraphic units. Axes are shared with Figs.~\ref{fig:fewshot_tiers} and~\ref{fig:fewshot_heads}.}
    \label{fig:fewshot_strongest}
\end{figure}

\paragraph{Fitting protocol.} Because rocks are collected one eruption at a time, we select whole stratigraphic units -- which contain numerous rocks per unit -- from the \rrTrainNRocks-rock fitting set. We add units at random up to sizes of \rsBudgetGrid{} rocks, with realized counts at or slightly above each size because units are sampled whole. At each size we make \fsBudgetDraws{} independent selections (replicates). For each replicate, the recalibration tiers of Appendix~\ref{sec:recalibrator_appendix} are fitted, and the three heads are trained on measured abundances (described below). Every fit is evaluated on the fixed validation set of \rrValNRocks{} rocks spanning \rrValNUnits{} stratigraphic units. The resulting fits are compared three ways: among the tiers (Fig.~\ref{fig:fewshot_tiers}), among the heads (Fig.~\ref{fig:fewshot_heads}), and between the mixture tier and the head given the latent representation (Fig.~\ref{fig:fewshot_strongest}).

\paragraph{Heads trained on measured abundances.} Each of the three heads follows the factorization of Eq.~\ref{eq:hurdle} over the \sysRealGroups{} mineral categories in the benchmark. Specifically, each factor of Eq.~\ref{eq:hurdle} is predicted by a multilayer perceptron with a single hidden layer of \rsHidden{} units. The abundance factor is the same \rsComponents-component Dirichlet mixture employed by the recalibration layer. The three heads differ only in their input, which is the frozen encoder's \sysEncoderDModel-dimensional latent representation, the composition the latent was computed from, or a zero vector. Compositions are converted to $\log_{10}$ concentration, with unreported elements set to zero. The input widths determine the parameter counts, where the head given the latent representation has \sysResupervisedParams{} parameters while the head given raw geochemistry has \sysResupervisedAssayParams{} parameters. 

As with the recalibrated model, all three heads minimize the joint NLL. Fitting employs full-batch Adam at a learning rate of \rsLr{} for \rsEpochs{} steps with no early stopping. The two values were determined from the training loss, where we took the largest learning rate under which the loss remains monotonic and the step count at which the loss plateaus (to within 0.01 nats of the final loss). The weight decay is chosen by the same protocol described in Appendix~\ref{sec:recalibrator_appendix}, where all three heads selected \rsWeightDecay{}. 

\paragraph{Head given no chemical inputs.} The head given no chemistry is introduced as a control, trained and evaluated exactly as the other heads are. Because the zero vector carries no chemical information, the presence factor can learn only the rate at which each mineral category appears across the rocks of a fit. The abundance factor, which conditions on the measured mineral list, is still learned as in the other two heads. At \fsRocksKnee{} rocks, the abundance-factor NLL of the head given the latent representation reaches \fsLatentAmountKnee{} nats compared to \fsAssemblageAmountKnee{} for the head given no chemistry and \fsAssayAmountKnee{} for the head given raw geochemistry, while at the full fitting set the three converge to \fsLatentAmountFull{}, \fsAssemblageAmountFull{}, and \fsAssayAmountFull{}. Presence prediction therefore benefits from chemical inputs. Meanwhile, abundance predictions depend primarily on the measured mineral list, benefiting only from chemical inputs when labeled rocks are scarce.

\paragraph{Comparing fits.} We report the contrast between two fits $a$ and $b$ over the replicates $r$ as:
\begin{equation}\label{eq:paired_contrast}
    \Delta_{ab} = \mathrm{median}_r\big(\mathrm{NLL}_a^{(r)} - \mathrm{NLL}_b^{(r)}\big)
\end{equation}
where the difference is taken within a replicate because both fits train on the same selected rocks. Values reported in Section~\ref{sec:results} are medians across the \fsBudgetDraws{} replicates. The head using the latent representation outperforms the head given raw geochemistry by \fsLatentOverAssayKnee{} nats of joint NLL at \sysFewshotBudgetKnee{} rocks (in \fsLatentOverAssayKneeDraws{} replicates) and by \fsLatentOverAssayFull{} nats at the full fitting set (in \fsLatentOverAssayFullDraws{} replicates). The head using the latent representation also outperforms the head given no chemistry by \fsLatentOverAssemblageKnee{} nats at \sysFewshotBudgetKnee{} rocks (in \fsLatentOverAssemblageKneeDraws{} replicates) and by \fsLatentOverAssemblageFull{} nats at the full fitting set (in \fsLatentOverAssemblageFullDraws{} replicates). Lastly, compared to the recalibration layer, the head given the latent representation is lower by a median of \fsLatentOverRecalKnee{} nats at \fsRocksKnee{} rocks and by \fsLatentOverRecalFull{} nats at the full fitting set. Thus, a head trained directly on measured abundances outperforms the recalibration layer in NLL (Fig.~\ref{fig:fewshot_strongest}a). However, the recalibration layer exhibits lower absolute error in every replicate from \fsRecalMaeUnanimousAt{} rocks onward (Fig.~\ref{fig:fewshot_strongest}b). We therefore observe a tradeoff between model confidence and model accuracy. We attribute the recalibrated model's weaker confidence to the assumed distribution of formation conditions (Eq.~\ref{eq:proposal}), which we sampled broadly. The spread of the predicted distributions therefore follows the sampling rather than a measurement. In contrast, a head trained on measured abundances does not inherit priors from the simulation. We posit that more physically-aligned formation conditions would tighten simulated labels, help remove the spread that recalibration is only capable of partly absorbing, and therefore improve model confidence.

\clearpage
\section{Supplementary Tables}
\renewcommand{\thetable}{E\arabic{table}}
\renewcommand{\theHtable}{E\arabic{table}}
\setcounter{table}{0}

\begin{table}[h]
\caption{Per-element imputation performance of the pretrained encoder. To score an element, its value is hidden from the encoder and a decoder recovers it from the frozen pooled latent representation. Each entry is the log-space $R^2$ between recovered and measured concentrations over \encImputeTestN{} held-out compositions. Cr occupies the ninth major-element row of the input vocabulary but is never reported in the pretraining compilation. Its row therefore always enters as padding, and no measured value exists to score against.}
\label{tab:imputation_per_element}
\centering
\small
\begin{tabular}{lccccccccccccc}
\toprule
\textbf{Major} & Si & Al & Ca & Mg & Fe & K & Na & Ti & Cr & & & & \\
$R^2$ & \encImputeRSqSi{} & \encImputeRSqAl{} & \encImputeRSqCa{} & \encImputeRSqMg{} & \encImputeRSqFe{} & \encImputeRSqK{} & \encImputeRSqNa{} & \encImputeRSqTi{} & -- & & & & \\
\midrule
\textbf{Trace} & Sr & Zr & Ba & Y & Rb & Ni & Nb & V & La & Ce & Nd & Yb & Th \\
$R^2$ & \encImputeRSqSr{} & \encImputeRSqZr{} & \encImputeRSqBa{} & \encImputeRSqY{} & \encImputeRSqRb{} & \encImputeRSqNi{} & \encImputeRSqNb{} & \encImputeRSqV{} & \encImputeRSqLa{} & \encImputeRSqCe{} & \encImputeRSqNd{} & \encImputeRSqYb{} & \encImputeRSqTh{} \\
\bottomrule
\end{tabular}
\end{table}
\vspace{3em}

\begin{table}[h]
\caption{Zero-shot presence discrimination per mineral category. AUROC of the predicted presence probability, computed over every rock in the benchmark.}
\label{tab:percat_auroc}
\centering
\small
\setlength{\tabcolsep}{4pt}
\begin{tabular}{lccccccc}
\toprule
& \multicolumn{1}{c}{\textbf{Feldspar}} &
\multicolumn{1}{c}{\textbf{Pyroxene}} &
\multicolumn{1}{c}{\textbf{Amphibole}} &
\multicolumn{1}{c}{\textbf{Biotite}} &
\multicolumn{1}{c}{\textbf{Olivine}} &
\multicolumn{1}{c}{\textbf{Fe-Ti oxides}} &
\multicolumn{1}{c}{\textbf{Quartz}} \\
\midrule
AUROC & \rrAurocFsp{} & \rrAurocPx{} & \rrAurocAmp{} & \rrAurocBi{} & \rrAurocOl{} & \rrAurocOpq{} & \rrAurocQ{} \\
\bottomrule
\end{tabular}
\end{table}
\vspace{3em}

\begin{table}[h]
\caption{Zero-shot abundance ranking per mineral category. Spearman rank correlation between predicted and measured abundance, computed over the $n$ rocks that report a quantity for each category.}
\label{tab:percat_spearman}
\centering
\small
\setlength{\tabcolsep}{4pt}
\begin{tabular}{lccccccc}
\toprule
& \multicolumn{1}{c}{\textbf{Feldspar}} &
\multicolumn{1}{c}{\textbf{Pyroxene}} &
\multicolumn{1}{c}{\textbf{Amphibole}} &
\multicolumn{1}{c}{\textbf{Biotite}} &
\multicolumn{1}{c}{\textbf{Olivine}} &
\multicolumn{1}{c}{\textbf{Fe-Ti oxides}} &
\multicolumn{1}{c}{\textbf{Quartz}} \\
\midrule
Spearman $\rho$ & \rrSpearmanFsp{} & \rrSpearmanPx{} & \rrSpearmanAmp{} & \rrSpearmanBi{} & \rrSpearmanOl{} & \rrSpearmanOpq{} & \rrSpearmanQ{} \\
$n$ & \rrNQuantFsp{} & \rrNQuantPx{} & \rrNQuantAmp{} & \rrNQuantBi{} & \rrNQuantOl{} & \rrNQuantOpq{} & \rrNQuantQ{} \\
\bottomrule
\end{tabular}
\end{table}
\vspace{3em}

\begin{table}[H]
\caption{Zero-shot abundance bias per mineral category. The percentage of rocks where predicted abundance exceeds the measured abundance; the median across rocks of the predicted-to-measured abundance ratio; and the number of rocks $n$ that report a quantity for the category.}
\label{tab:percat_bias}
\centering
\small
\setlength{\tabcolsep}{4pt}
\begin{tabular}{lccccccc}
\toprule
& \multicolumn{1}{c}{\textbf{Feldspar}} &
\multicolumn{1}{c}{\textbf{Pyroxene}} &
\multicolumn{1}{c}{\textbf{Amphibole}} &
\multicolumn{1}{c}{\textbf{Biotite}} &
\multicolumn{1}{c}{\textbf{Olivine}} &
\multicolumn{1}{c}{\textbf{Fe-Ti oxides}} &
\multicolumn{1}{c}{\textbf{Quartz}} \\
\midrule
Rocks over-predicted (\%) & \rrOverPctFsp{} & \rrOverPctPx{} & \rrOverPctAmp{} & \rrOverPctBi{} & \rrOverPctOl{} & \rrOverPctOpq{} & \rrOverPctQ{} \\
Median pred/measured & \rrMagFoldFsp{} & \rrMagFoldPx{} & \rrMagFoldAmp{} & \rrMagFoldBi{} & \rrMagFoldOl{} & \rrMagFoldOpq{} & \rrMagFoldQ{} \\
$n$ & \rrNQuantFsp{} & \rrNQuantPx{} & \rrNQuantAmp{} & \rrNQuantBi{} & \rrNQuantOl{} & \rrNQuantOpq{} & \rrNQuantQ{} \\
\bottomrule
\end{tabular}
\end{table}

\newpage
\section{Supplementary Figures}
\renewcommand{\thefigure}{F\arabic{figure}}
\renewcommand{\theHfigure}{F\arabic{figure}}
\setcounter{figure}{0}

\begin{figure}[h]
    \centering
    \includegraphics[width=1.0\linewidth]{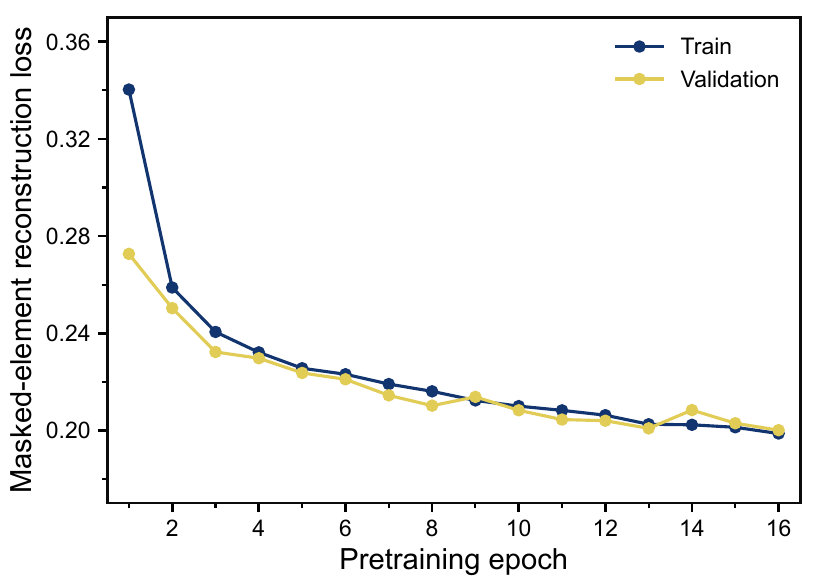}
    \caption{\textbf{Composition encoder pretraining loss.} Masked-element reconstruction loss (dimensionless) over pretraining of the composition encoder, where the loss is the standardized log MSE as presented in Eq.~\ref{eq:ssl}. The validation loss is evaluated on held-out compositions from the pretraining set. Training was halted by early stopping once the validation ceased to improve 0.001/epoch. }
    \label{fig:pretrain_loss}
\end{figure}
\clearpage

\begin{figure}[h]
    \centering
    \includegraphics[width=1.0\linewidth]{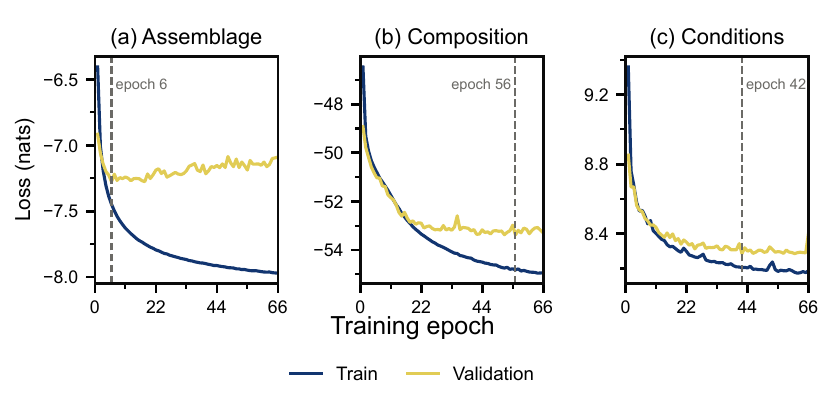}
    \caption{\textbf{Per-head training loss.} Loss over joint-heads training, where each head is saved at its own best-validation epoch (vertical dashed line). The validation loss is evaluated on the \sysCorpusValRows{} rows of the validation compositions (Table~\ref{tab:corpus_scale}). \textbf{(a)} Assemblage head, the summed presence and abundance loss. \textbf{(b)} Composition head. \textbf{(c)} Conditions head.}
    \label{fig:heads_loss}
\end{figure}
\clearpage

\begin{figure}[h]
    \centering
    \includegraphics[width=1.0\linewidth]{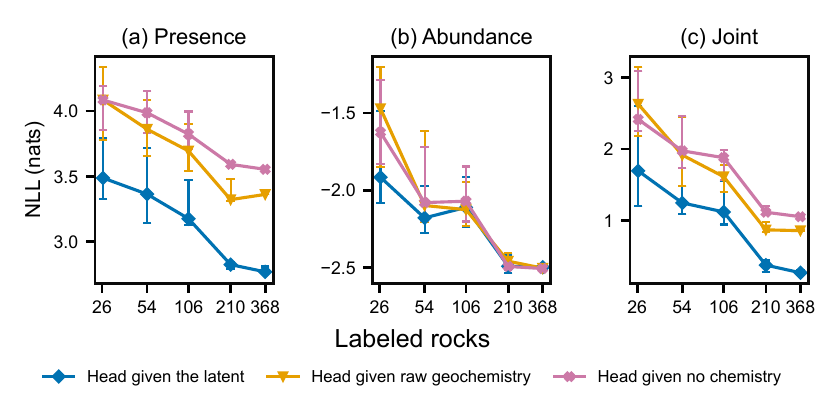}
    \caption{\textbf{Joint NLL decomposed into presence and abundance factors.} The three heads trained on measured abundances, differing only in their input (Appendix~\ref{sec:resupervision_appendix}). \textbf{(a)} The presence (Bernoulli) factor of Eq.~\ref{eq:hurdle}. \textbf{(b)} The abundance (Dirichlet) factor, where all three heads are given the measured mineral list as their support (i.e., teacher-forced). \textbf{(c)} The joint NLL, the sum of the presence and abundance factors. Note that this panel corresponds exactly to Fig.~\ref{fig:fewshot_heads}a. Points are medians and error bars span the interquartile range over the \fsBudgetDraws{} replicates at each fitting size, evaluated on the benchmark validation set. Note that the gap between curves is a difference of medians because each curve summarizes one head separately.}
    \label{fig:fewshot_decomp}
\end{figure}
\clearpage

\begin{figure}[h]
    \centering
    \includegraphics[width=1.0\linewidth]{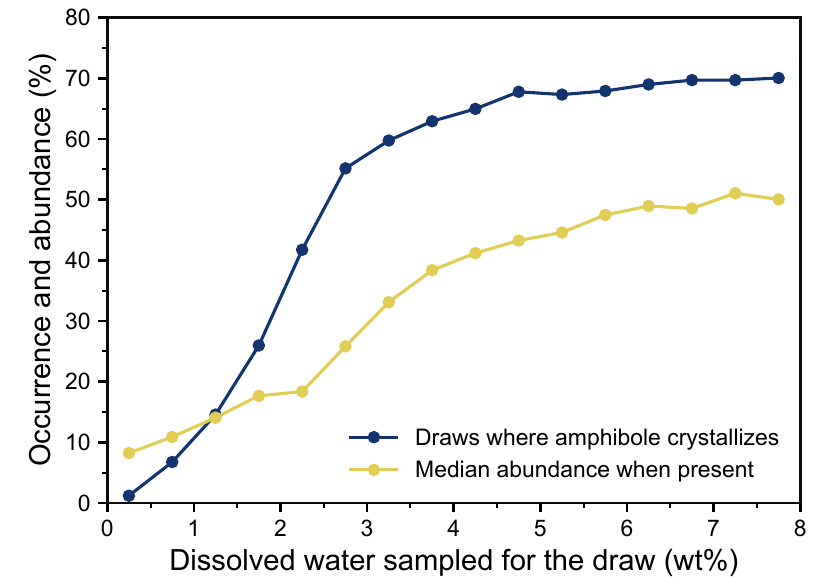} 
    \caption{\textbf{Water sensitivity of amphibole crystallization.} Amphibole occurrence and abundance in the simulated label corpus, pooled over the validation compositions.}
    \label{fig:amp_water_dose} 
\end{figure}
\clearpage


\newpage
\section*{NeurIPS Paper Checklist}

\begin{enumerate}

\item {\bf Claims}
    \item[] Question: Do the main claims made in the abstract and introduction accurately reflect the paper's contributions and scope?
    \item[] Answer: \answerYes{} 
    \item[] Justification: All main claims are supported with figures and tables in the main text and appendices. In addition, the introduction explicitly scopes the model's capabilities to the equilibrium crystallization of volcanic rocks. 

\item {\bf Limitations}
    \item[] Question: Does the paper discuss the limitations of the work performed by the authors?
    \item[] Answer: \answerYes{} 
    \item[] Justification: Limitations are discussed throughout the paper as they arise, including the thermodynamic simulators' omissions of kinetics, transport, and measured formation conditions (Section~\ref{sec:intro}); the scope of the model to equilibrium crystallization of volcanic rocks (Section~\ref{sec:intro}); the systematic abundance bias inherited from the simulated labels (Fig.~\ref{fig:cohort_card}c; Table~\ref{tab:percat_bias}); and the methodological assumptions over formation conditions, which impacts the spread of the predicted distributions (Eq.~\ref{eq:proposal} and Appendix~\ref{sec:resupervision_appendix}).

\item {\bf Theory assumptions and proofs}
    \item[] Question: For each theoretical result, does the paper provide the full set of assumptions and a complete (and correct) proof?
    \item[] Answer: \answerNA{} 
    \item[] Justification: The paper is empirical and presents no theorems or proofs.

    \item {\bf Experimental result reproducibility}
    \item[] Question: Does the paper fully disclose all the information needed to reproduce the main experimental results of the paper to the extent that it affects the main claims and/or conclusions of the paper (regardless of whether the code and data are provided or not)?
    \item[] Answer: \answerYes{} 
    \item[] Justification: Every main claim made in the paper has an experimental setup described in detail (Appendix~\ref{sec:implementation_appendix}), including tokenization, encoder configuration, pretraining, simulated-corpus generation, head architectures, joint training, recalibration fitting, and benchmark construction. All data splits, hyperparameter selection procedures, and evaluation protocols are detailed in a manner that allows a reader to reproduce the work. 

\item {\bf Open access to data and code}
    \item[] Question: Does the paper provide open access to the data and code, with sufficient instructions to faithfully reproduce the main experimental results, as described in supplemental material?
    \item[] Answer: \answerYes{} 
    \item[] Justification: Code, configuration files, a version-pinned environment specification, both trained model checkpoints (encoder and prediction heads), the benchmark tables, and the figure pipeline are available at \url{https://github.com/matthew-j-liu/phasecast}. 

\item {\bf Experimental setting/details}
    \item[] Question: Does the paper specify all the training and test details (e.g., data splits, hyperparameters, how they were chosen, type of optimizer) necessary to understand the results?
    \item[] Answer: \answerYes{} 
    \item[] Justification: The optimizer, learning rate, and schedule are specified in every training stage of the model architecture. In addition, hyperparameter values, search spaces, and selection criteria are tabulated in Tables~\ref{tab:encoder_hparams}, \ref{tab:pretrain_hparams}, and~\ref{tab:heads_hpo}. All data splits and evaluation protocols are defined in Section~\ref{sec:methods} and Appendix~\ref{sec:implementation_appendix}.

\item {\bf Experiment statistical significance}
    \item[] Question: Does the paper report error bars suitably and correctly defined or other appropriate information about the statistical significance of the experiments?
    \item[] Answer: \answerYes{} 
    \item[] Justification: The model presented in this work predicts distributions, which we always show the spread for (e.g., whiskers spanning the interquartile range). Other sources of uncertainty are described as they arise throughout the paper, including benchmark variability by bootstrap resampling (Appendix~\ref{sec:discrimination_appendix}) and training variability due to the random selection of training rocks (Appendix~\ref{sec:resupervision_appendix}).

\item {\bf Experiments compute resources}
    \item[] Question: For each experiment, does the paper provide sufficient information on the computer resources (type of compute workers, memory, time of execution) needed to reproduce the experiments?
    \item[] Answer: \answerYes{}
    \item[] Justification: A paragraph at the beginning of Appendix~\ref{sec:implementation_appendix} details the hardware and time of execution needed for all experiments (e.g., corpus generation, encoder pretraining, hyperparameter searches, head training, and recalibration). In addition, an approximate compute time for unreported, preliminary iterations is provided. 
    
\item {\bf Code of ethics}
    \item[] Question: Does the research conducted in the paper conform, in every respect, with the NeurIPS Code of Ethics \url{https://neurips.cc/public/EthicsGuidelines}?
    \item[] Answer: \answerYes{} 
    \item[] Justification: Our work uses publicly available scientific data and does not involve human subjects, personal data, or other activities that raise ethical concerns under the NeurIPS Code of Ethics.

\item {\bf Broader impacts}
    \item[] Question: Does the paper discuss both potential positive societal impacts and negative societal impacts of the work performed?
    \item[] Answer: \answerYes{} 
    \item[] Justification: Positive impacts such as higher-throughput characterization of modal mineralogy for geology are discussed in Section~\ref{sec:intro}. We foresee no direct path to malicious applications from predicting rock mineralogy, though negative effects for a user of the model might include a downstream decision predicated on potentially inaccurate zero-shot predictions. The paper addresses this concern by prescribing recalibration on measured rocks.
    
\item {\bf Safeguards}
    \item[] Question: Does the paper describe safeguards that have been put in place for responsible release of data or models that have a high risk for misuse (e.g., pre-trained language models, image generators, or scraped datasets)?
    \item[] Answer: \answerNA{} 
    \item[] Justification: Neither the model, which predicts rock mineralogy, nor the data (from published geochemical and petrographic sources) carries a high risk for misuse.

\item {\bf Licenses for existing assets}
    \item[] Question: Are the creators or original owners of assets (e.g., code, data, models), used in the paper, properly credited and are the license and terms of use explicitly mentioned and properly respected?
    \item[] Answer: \answerYes{} 
    \item[] Justification: All datasets and software are cited in the references. Their licenses permit our use: the geochemical compilation is distributed CC BY 4.0, the two USGS petrographic studies are public-domain U.S. Government works, \texttt{MAGEMin} is GPL-3.0 and is called as an unmodified solver, and CrabNet, the mat2vec embeddings, and Optuna are MIT-licensed.

\item {\bf New assets}
    \item[] Question: Are new assets introduced in the paper well documented and is the documentation provided alongside the assets?
    \item[] Answer: \answerYes{} 
    \item[] Justification: The new assets (trained encoder and head weights, the compiled benchmark tables, and the simulated label corpus) are described in Appendices~\ref{sec:architecture_appendix}, \ref{sec:pretrain_appendix}, \ref{sec:heads_appendix}, and \ref{sec:training_appendix} (the weights), Appendix~\ref{sec:benchmark_appendix} (the benchmark tables), and Appendix~\ref{sec:corpus_appendix} (the label corpus). In addition, the repository details the codebase and provides a version-pinned environment; due to size constraints, the repository carries an excerpt of the corpus and links to the full set, while Appendix~\ref{sec:corpus_appendix} gives the full generation procedure.
    

\item {\bf Crowdsourcing and research with human subjects}
    \item[] Question: For crowdsourcing experiments and research with human subjects, does the paper include the full text of instructions given to participants and screenshots, if applicable, as well as details about compensation (if any)? 
    \item[] Answer: \answerNA{} 
    \item[] Justification: The paper involves no crowdsourcing and no research with human subjects. All data are physical measurements of rocks from published sources.

\item {\bf Institutional review board (IRB) approvals or equivalent for research with human subjects}
    \item[] Question: Does the paper describe potential risks incurred by study participants, whether such risks were disclosed to the subjects, and whether Institutional Review Board (IRB) approvals (or an equivalent approval/review based on the requirements of your country or institution) were obtained?
    \item[] Answer: \answerNA{} 
    \item[] Justification: The paper involves no research with human subjects, so IRB or equivalent review is not applicable.

\item {\bf Declaration of LLM usage}
    \item[] Question: Does the paper describe the usage of LLMs if it is an important, original, or non-standard component of the core methods in this research? Note that if the LLM is used only for writing, editing, or formatting purposes and does \emph{not} impact the core methodology, scientific rigor, or originality of the research, declaration is not required.
    \item[] Answer: \answerNA{} 
    \item[] Justification: LLM tools assisted with writing and code implementation as general-purpose assistants, but were not a component of the core methodology or scientific reasoning in this work. The motivation, framing, and all primary design decisions were made by the authors, guided by their own expertise and intuition.

\end{enumerate}

\end{document}